\documentclass[10pt,twocolumn,letterpaper]{article}

\usepackage{cvpr}              % To produce the CAMERA-READY version
\definecolor{cvprblue}{rgb}{0.21,0.49,0.74}
\usepackage[pagebackref,breaklinks,colorlinks,allcolors=cvprblue]{hyperref}

\usepackage{multirow}
\usepackage{adjustbox} % Add this line
\usepackage{multirow} % Add this line at the beginning of your LaTe
\usepackage{makecell}
\usepackage{pifont}

\usepackage[utf8]{inputenc} % allow utf-8 input
\usepackage[T1]{fontenc}    % use 8-bit T1 fonts
\usepackage{hyperref}       % hyperlinks
\usepackage{url}            % simple URL typesetting
\usepackage{booktabs}       % professional-quality tables
\usepackage{amsfonts}       % blackboard math symbols
\usepackage{nicefrac}       % compact symbols for 1/2, etc.
\usepackage{microtype}      % microtypography
\usepackage{xcolor}         % colors
\usepackage{graphicx} % for \includegraphics
\usepackage{subcaption} % for subfigure environment
\usepackage{wrapfig} % for wraptable environment
\usepackage{booktabs} % for \toprule, \midrule, \bottomrule
\usepackage{caption} % for \captionsetup

\usepackage{pifont}
\usepackage[dvipsnames]{xcolor}
\usepackage{colortbl}

\usepackage{graphicx}
\usepackage{booktabs}

\def\confName{CVPR}
\def\confYear{2025}

\title{Repurposing Pre-trained Video Diffusion Models \\ for Event-based Video Interpolation}

\author{Jingxi Chen$^{1}$, Brandon Y. Feng$^{2}$, Haoming Cai$^{1}$, Tianfu Wang$^{1}$, Levi Burner$^{1}$, Dehao Yuan$^{1}$, \\ Cornelia Ferm{\"u}ller$^{1}$,  Christopher A. Metzler$^{1}$, Yiannis Aloimonos$^{1}$\\
$^{1}$University of Maryland, College Park \ \ $^{2}$Massachusetts Institute of Technology\\
\url{https://vdm-evfi.github.io/}
}

\begin{document}

\twocolumn[{%
\renewcommand\twocolumn[1][]{#1}%
\maketitle
\begin{center}
    \captionsetup{type=figure}
    \includegraphics[scale=0.58]{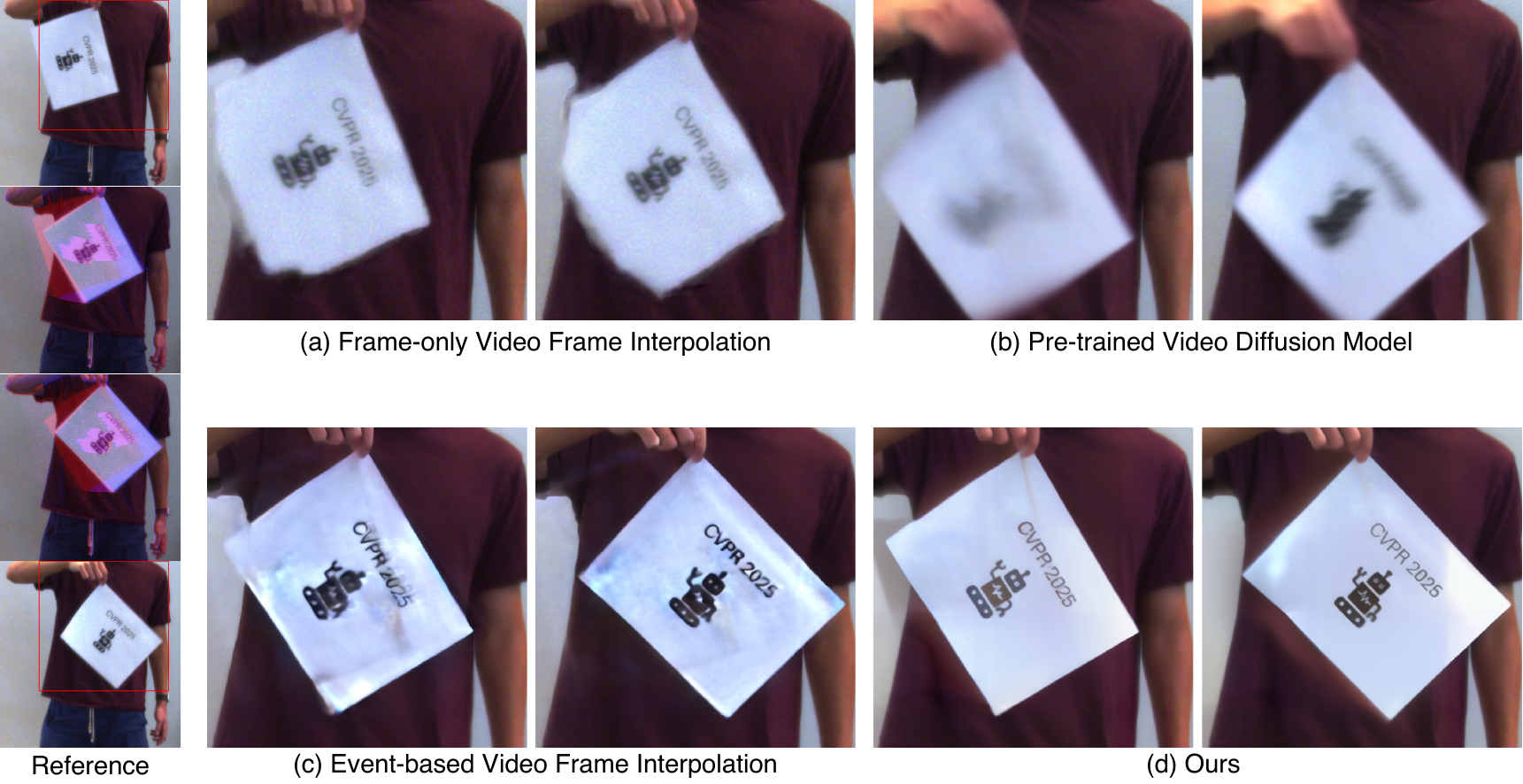} % Adjust the scale as needed
    \vspace{-5pt}
    \captionsetup{font=small}
    \caption{We compare our proposed approach RE-VDM, which adapts a pre-trained video diffusion model for event-based video frame interpolation on unseen real-world data, with three baselines: frame-only interpolation method GIMM-VFI-R-P \cite{guo2024generalizable}, test-time optimization of a pre-trained video diffusion model via Time Reversal \cite{feng2025explorative}, and event-based interpolation method CBMNet-Large \cite{Kim_2023_CVPR}, trained on the same dataset as our method. The left-most column shows the start frame, end frame, and reference interpolation frames overlaid with events. Leveraging data priors in the pre-trained video diffusion model and the diffusion process, along with motion guidance controlled by input events, our approach demonstrates superior generalization performance on unseen real-world frames with substantial motion.}
    \label{fig:teaser}
    \vspace{-3pt}
\end{center}%
}]

\vspace{-10pt}
\begin{abstract}

Video Frame Interpolation aims to recover realistic missing frames between observed frames, generating a high-frame-rate video from a low-frame-rate video. However, without additional guidance, the large motion between frames makes this problem ill-posed. Event-based Video Frame Interpolation (EVFI) addresses this challenge by using sparse, high-temporal-resolution event measurements as motion guidance. This guidance allows EVFI methods to significantly outperform frame-only methods. However, to date, EVFI methods have relied on a limited set of paired event-frame training data, severely limiting their performance and generalization capabilities. 
In this work,  we overcome the limited data challenge by adapting pre-trained video diffusion models trained on internet-scale datasets to EVFI. 
We experimentally validate our approach on real-world EVFI datasets, including a new one that we introduce. Our method outperforms existing methods and generalizes across cameras far better than existing approaches.

\end{abstract}    
\section{Introduction}
\label{sec:intro}

Event cameras are a novel class of neuromorphic sensors offering unique advantages, including high dynamic range and high temporal resolution \cite{gallego2020event}. One significant application of event cameras is Event-based Video Frame Interpolation (EVFI) \cite{tulyakov2021time, tulyakov2022time, zhang2022unifying, sun2023event, ma2025timelens, Kim_2023_CVPR}. By capturing traditional frames alongside high temporal resolution events, event data can fill in missing motion information between frames, aiding in interpolation. This approach avoids the ill-posed problem of frame-only interpolation, where large motion between frames introduces excessive degrees of freedom, making the interpolation of frames infeasible.

Current state-of-the-art (SOTA) EVFI methods achieve impressive performance when trained and tested on the same dataset. However, when the test set differs from the training set with changes in data and motion distribution, these methods experience significant performance degradation, as we can see in Fig. \ref{fig:teaser}. This issue arises from two main factors. First, compared to broader fields like video generation, the EVFI field is much smaller, with limited data quantity and quality. Second, the models used in EVFI are highly specialized, with constrained representational power.

% in parallel Generative AI utilizes massive (tranistion misisng)
Recent advances in generative AI, particularly in video generation, have spurred significant progress in the field. Leading companies have invested substantial resources into building massive, high-quality datasets \cite{blattmann2023align} and developing foundation models for video generation \cite{ho2022video, blattmann2023align}. Most of these foundation models are video diffusion models. By leveraging millions of high-quality commercial video clips and billions of parameters, these diffusion models achieve impressive generalization performance compared to previous approaches. A comparison between video generation and EVFI fields is presented in Table \ref{tab:compare_fields}. The primary differences between these fields lie in dataset quality/size and model parameter scale, the EVFI field lacks the extensive commercial datasets and large model sizes found in video generation, impacting its models' generalization and representational power.

This raises a natural question: 
\textit{Can we adapt pre-trained video diffusion foundation models for EVFI to leverage their learned data priors and model design advantages?}

In this work, we introduce the first approach, RE-VDM, for adapting pre-trained video diffusion foundation models to the EVFI task, aiming to bridge EVFI with generative AI by leveraging robust data priors and foundation models. To achieve this, we address key technical challenges. First, for data-efficient adaptation, we train an additional subset of the model with event-based control on small EVFI datasets inspired by \cite{zhang2023adding}, while keeping the original weights frozen to avoid catastrophic forgetting of learned data priors as explained in Section \ref{data_efficient}. Second, while video generation aims to produce diverse videos without strict realism constraints, EVFI requires ground-truth fidelity to accurately recover missing frames. To preserve motion and appearance fidelity, we employ upsampling and Per-tile Denoising and Fusion as explained in Section \ref{sec:overlap_tile} as the test-time optimization for video generation with high-fidelity appearance and event-based motion control. Finally, unlike video generation, which is an extrapolation task based on an image or text prompt, EVFI requires interpolation between the start and end frames. To address this, we develop a test-time optimization approach that generates event-based video latents from both the start and end frames at each denoising step, fusing them for a consistent, interpolated result that incorporates information from both frames, as explained in Section \ref{sec:interpolation}

For a robust evaluation, we compare the generalization performance of our approach against representative frame-only, event-based, and pre-trained video diffusion models for frame interpolation across multiple datasets, including self-collected Clear-Motion test sequences. The results demonstrate the potential and advantages of this adaptation. We also discuss the limitations observed, highlighting the exploratory nature of our work. To summarize, our contributions are 
\begin{itemize}
    \item We present the first approach for adapting pre-trained video diffusion foundation models to the Event-based Video Frame Interpolation task, leveraging the strong data priors learned from large video generation datasets and the inherent advantages of video diffusion models.
    \item Experimental results demonstrate the strong potential of the adapted video diffusion model for the EVFI task, showing excellent generalization on unseen real-world data and consistency in reconstructed frames.
\end{itemize}

\begin{table}[t!]
    \centering
    \resizebox{1.0\linewidth}{!}{
    \begin{tabular}{@{}cccc@{}} % Four columns
    \midrule
    \midrule
    & \makecell{\textbf{Feature}} & \makecell{\textbf{Video} \\ \textbf{Generation}} & \makecell{\textbf{EVFI}} \\
    \midrule
     & Dataset Size \cite{blattmann2023stable, polyak2024movie, liu2024sora} & $10^{7}$ & $10^{2}$ \\
     & Dataset Quality \cite{blattmann2023stable, polyak2024movie, liu2024sora, Kim_2023_CVPR, tulyakov2022time, stoffregen2020reducing} & Commercial & Custom \\
     & Model Type \cite{blattmann2023stable, tulyakov2022time, sun2023event, ma2025timelens, Kim_2023_CVPR} & Video Diffusion (Foundation) Models & Custom Models \\
     & Model Size \cite{blattmann2023stable, polyak2024movie, tulyakov2022time, sun2023event, ma2025timelens, Kim_2023_CVPR} & $10^{10}$ & $10^{8}$ \\
     & Video Generation Type \cite{blattmann2023stable, tulyakov2022time, sun2023event, ma2025timelens, Kim_2023_CVPR} & Extrapolation & Interpolation \\
    \midrule
    \midrule
    \end{tabular}}

    \captionsetup{font=small}
    \caption{Comparison between video generation and event-based video frame interpolation (EVFI) fields reveals major differences in dataset size, quality, and model design. Video generation datasets are orders of magnitude larger than EVFI datasets and are of commercial-level quality. Models for video generation are typically foundation models with billions of parameters. With abundant training data and large-scale foundation models, video diffusion models are well-suited for generalized video generation tasks. }
    \label{tab:compare_fields}

    \vspace{-6mm}
\end{table}

\section{Related Work}
\label{sec:related_work}

\subsection{Event Camera and Video Frame Interpolation}
Event cameras are neuromorphic sensors that capture the brightness changes in the scene, and the brightness changes are primarily due to the camera's motion and object motion in the scene. For this reason, events captured by event cameras can be used to compute motion-related information, for example, optical flow, motion segmentation, and ego-motion \cite{benosman2012asynchronous,barranco2014contour,bardow2016simultaneous, barranco2018real,mitrokhin2018event, he2024microsaccade, shah2024codedevents, zhu2019unsupervised, zhu2018ev, yuan2024learning}. Various methods have been proposed to reconstruct intensity videos from event streams alone, demonstrating that while feasible, the quality of such reconstructions is significantly limited by camera and object motion \cite{cadena2023sparse, rebecq2019high}. Combining events with images extends beyond event-only applications, offering powerful capabilities for tasks like image deblurring and video frame interpolation/extrapolation \cite{tulyakov2021time, tulyakov2022time, zhang2022unifying, chen2024timerewind, sun2023event}. This approach leverages events as cues to bridge the gap between frames, thus avoiding the ill-posed challenges faced by frame-only methods \cite{huang2022real, park2021asymmetric, lu2022video, jiang2018super, bao2019depth, kong2022ifrnet, Wu_2024_CVPR, zhong2024clearer, zhang2023extracting, guo2024generalizable} when interpolating between frames with substantial motion.

\subsection{Diffusion and Video Diffusion Models}

Diffusion models~\cite{ho2020denoising, song2020score} is an emerging class of image generative models that model the reverse diffusion process of recovering data from noise. 
Starting from data $\mathbf{x}_0$ from the data distribution $p_{data}(\mathbf{x})$, The diffusion model tries to reverse the forward diffusion process as an $N$ step Markov chain $\{\mathbf{x_i}\}, i \in [0, N]$ that satisfies the marginal distribution 
\begin{equation}
p_{\alpha_i}(\mathbf{x}_i | \mathbf{x}_0) = \mathcal{N}(\mathbf{x}_i; {\alpha(t)}\mathbf{x}_0, \sigma(t)^2\mathbf{I}).
\end{equation}
Here, $\alpha(t), \sigma(t)$ represents a noise schedule. 
Diffusion models have demonstrated remarkable image generation quality and diversity. Stable Diffusion~\cite{rombach2022high} 
shifts the generation process to a low-dimensional latent space, greatly reducing computational cost.
Trained on massive online text-image pair datasets~\cite{schuhmann2021laion}, Stable Diffusion exhibits strong image generation capabilities. Video diffusion models~\cite{ho2022video, blattmann2023align} add temporal layers on top of existing image diffusion models to jointly denoise multiple consistent image frames. Stable Video Diffusion (SVD)~\cite{blattmann2023align} is built on top of latent diffusion model and trained on massive video datasets, serving as a strong foundational model for video generation tasks.

\subsection{Controllable Diffusion Generation Through Fine-Tuning}
There have been numerous works on generation conditioned on additional user inputs or control~\cite{rombach2022high, ruiz2023dreambooth, zhang2023adding, xing2024dynamicrafter}.
Stable Diffusion~\cite{rombach2022high} uses cross-attention layers in the diffusion U-Net to inject text control.
Dreambooth~\cite{ruiz2023dreambooth} showed that a diffusion model can be fine-tuned for conditional control.
ControlNet~\cite{zhang2023adding} modifies a trainable copy of the original diffusion U-Net for conditional control.
In addition to fine-tuning diffusion models for added controllability, various training-free methods have been proposed to alter the sampling process of image and video diffusion models for more flexible and generalized content synthesis~\cite{bar2023multidiffusion, feng2025explorative, lugmayr2022repaint, zhang2024mimicmotion}.
Latent diffusion models have limitations on generating a limited set of resolutions in a downsampled latent space.
MultiDiffusion~\cite{bar2023multidiffusion} provides a solution for resolution-agnostic generation by tile-based denoising.
When combined with conditional control, MultiDiffusion~\cite{bar2023multidiffusion} has shown to be effective in generating synthetic detailed images under the condition of fine-grained condition control~\cite{jia2025dginstyle, wang2024consistency}. 
On the video generation side, ~\cite{feng2025explorative} proposes a test-time sampling method that empowers video diffusion models to control camera and object motion. ~\cite{zhang2024mimicmotion} enables consistent long-frame generation outside of the trained frame length of a video diffusion model.
\section{Proposed Approach}

\subsection{Pipeline Overview}
\begin{figure}[!t]
    \centering
    \includegraphics[width=.999\linewidth]{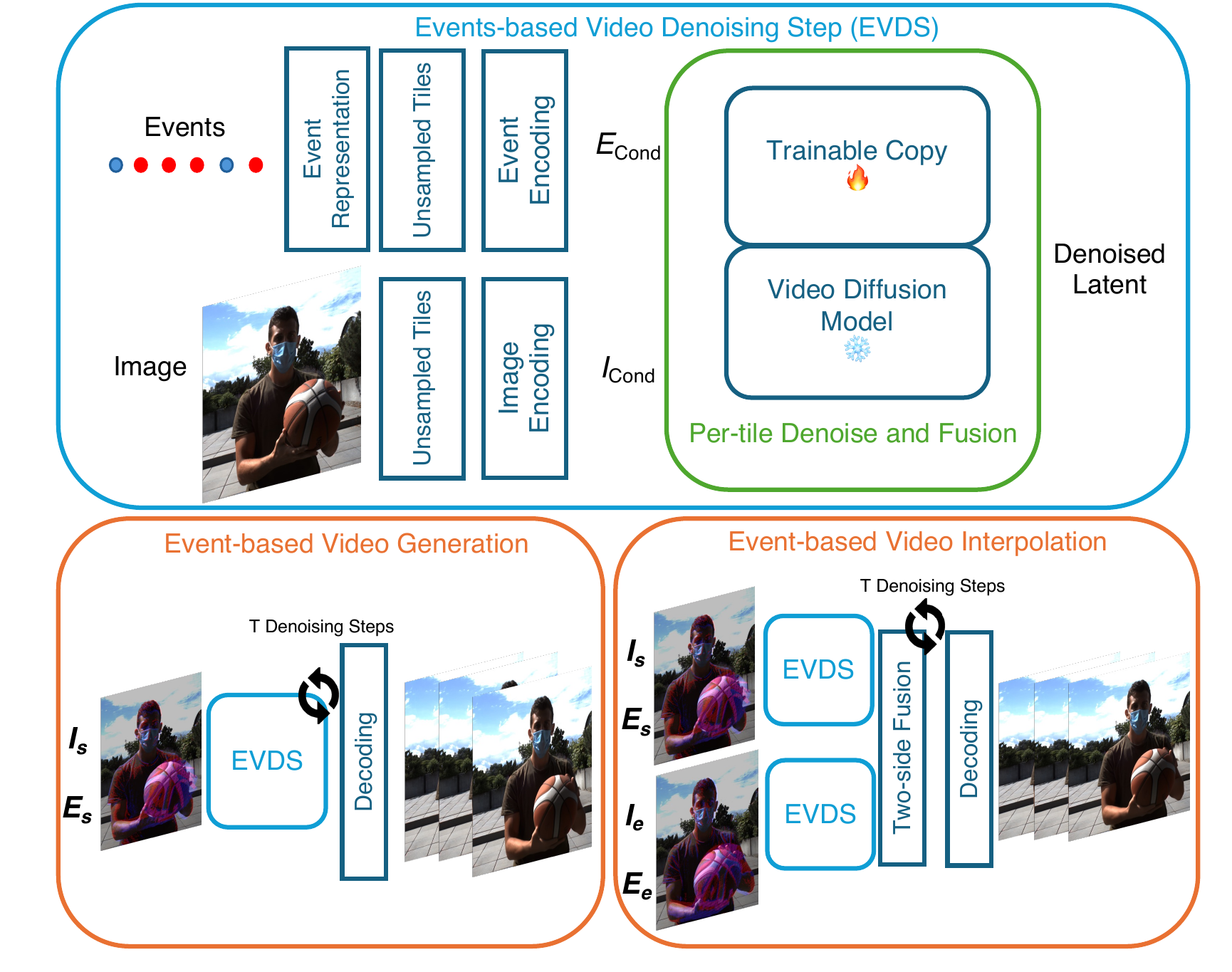}  

    \caption{The overview of our proposed approach RE-VDM: for adapting pre-trained video diffusion models includes two tasks: event-based video generation and interpolation. For video generation, the method utilizes the start frame $I_{s}$ and forward-time events $E_{s}$. For interpolation, it incorporates both the start frame $I_{s}$ and forward-time events $E_{s}$, as well as the end frame $I_{e}$ and backward-time events $E_{e}$ to achieve consistent results. Unlike video generation, for interpolation, a denoising step $t$ concludes with Two-side Fusion instead of EVDS.}

    \label{fig:pipe_overview}
    \vspace{-8pt}
\end{figure}

Figure \ref{fig:pipe_overview} presents an overview of our pipeline RE-VDM, for adapting pre-trained video diffusion models to event-based video generation and interpolation tasks. The key difference is that interpolation applies our EVDS on both the start frame $I_{s}$ for forward-time video denoising/generation and the end frame $I_{e}$ for backward-time video denosing/generation. At each denoising step, forward-backward consistency is enforced to achieve accurate interpolation. 
% Details and design of each key component are explained in the following.

In this work, without loss of generality, we assume that the pre-trained video diffusion model used is a Latent Diffusion Model, as is common in most video diffusion models.

\subsection{Data-Efficient Adaptation of Pre-trained Video Diffusion Models}
\label{data_efficient}
In this section we will explain the details for introducing the additional event-based motion control to pre-trained video diffusion models. As discussed in Section \ref{sec:intro} and shown in Table \ref{tab:compare_fields}, the large training datasets and model size of pre-trained video diffusion foundation models provide impressive generalization power for video generation tasks. However, these factors also limit the fine-tuning flexibility of the pre-trained model weights on existing EVFI datasets due to the risk of catastrophic forgetting, stemming from significant differences in data size and quality.

\begin{figure}[!t]
    \centering
    \includegraphics[width=.999\linewidth]{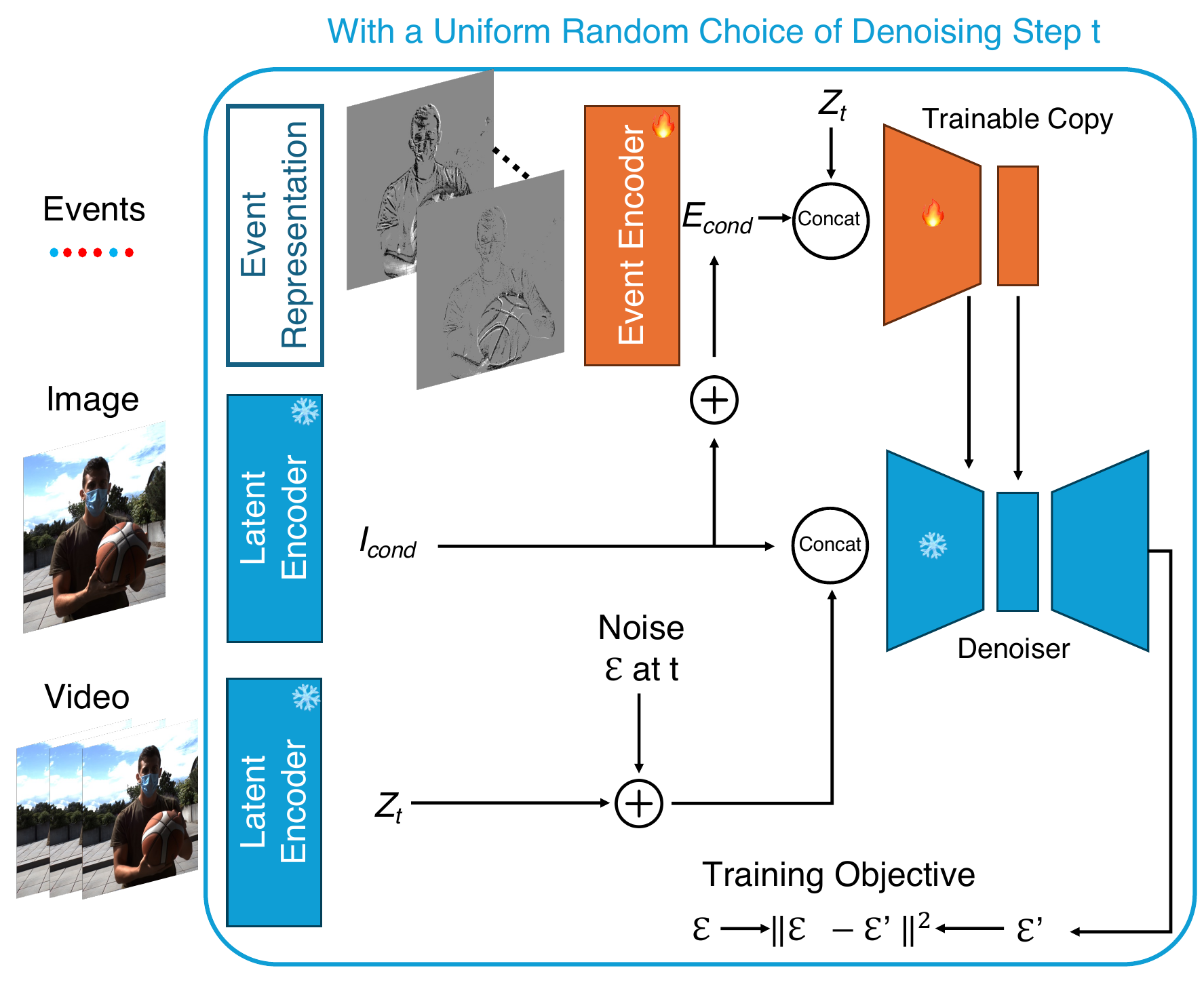}  

    \caption{The illustration depicts our training scheme for adapting a pre-trained video diffusion model to event-based video denoising. Our approach uses a frozen denoiser network from the pre-trained model, augmented with a trainable subset of blocks copied from the frozen denoiser. }

    \label{fig:EVDS_train}
    \vspace{-5pt}
\end{figure}

To prevent catastrophic forgetting of the learned data priors in pre-trained video diffusion foundation models, we drew inspiration from the \cite{zhang2023adding} approach for controllable image diffusion models, known for its impressive performance with small data adaptation. In our design, we keep the original weights of the video diffusion models frozen during training. Event-based control is introduced by training only a subset of blocks, copied from the original denoiser network, to serve as additional residuals to corresponding block outputs in the frozen denoiser network. A detailed illustration is provided in Figure \ref{fig:EVDS_train}.

The added event encoder and trainable copy aim to learn the correct encoding for events as the control condition, $E_{cond}$, as well as the block output residuals for the frozen pre-trained denoiser network by optimizing the trainable copy. We use the standard diffusion objective, which minimizes the mean squared error loss between the predicted noise $\epsilon'$ and the ground truth noise $\epsilon$ added to the video latent for a uniformly randomly sampled denoising time $t$.

During inference, the only difference from the training scheme is the absence of a given video latent $Z_{t}$ and noise $\epsilon$ for a sampled denoising time $t$. Instead, the initial noisy latent $Z_{T}$ is generated from standard Gaussian noise $N(0, 1)$ at the beginning of the denoising process.

% \begin{figure*}[!tbh]
%     \centering
%     \includegraphics[width=.999\linewidth]{sec/Figures/Pipelein_Details.pdf}  

%     \caption{Overview about the each components of the proposed pipeline and the relationship between each part. The detail of Overlapping Tile Fusion can be found in section \ref{sec:overlap_tile}, the detail of Two-side Fusion can be found in section \ref{sec:interpolation}}

%     \label{fig:pipe_details}
%     \vspace{-5pt}
% \end{figure*}

\subsection{Events as the Control Condition}
\begin{figure}[!t]
    \centering
    \includegraphics[width=.999\linewidth]{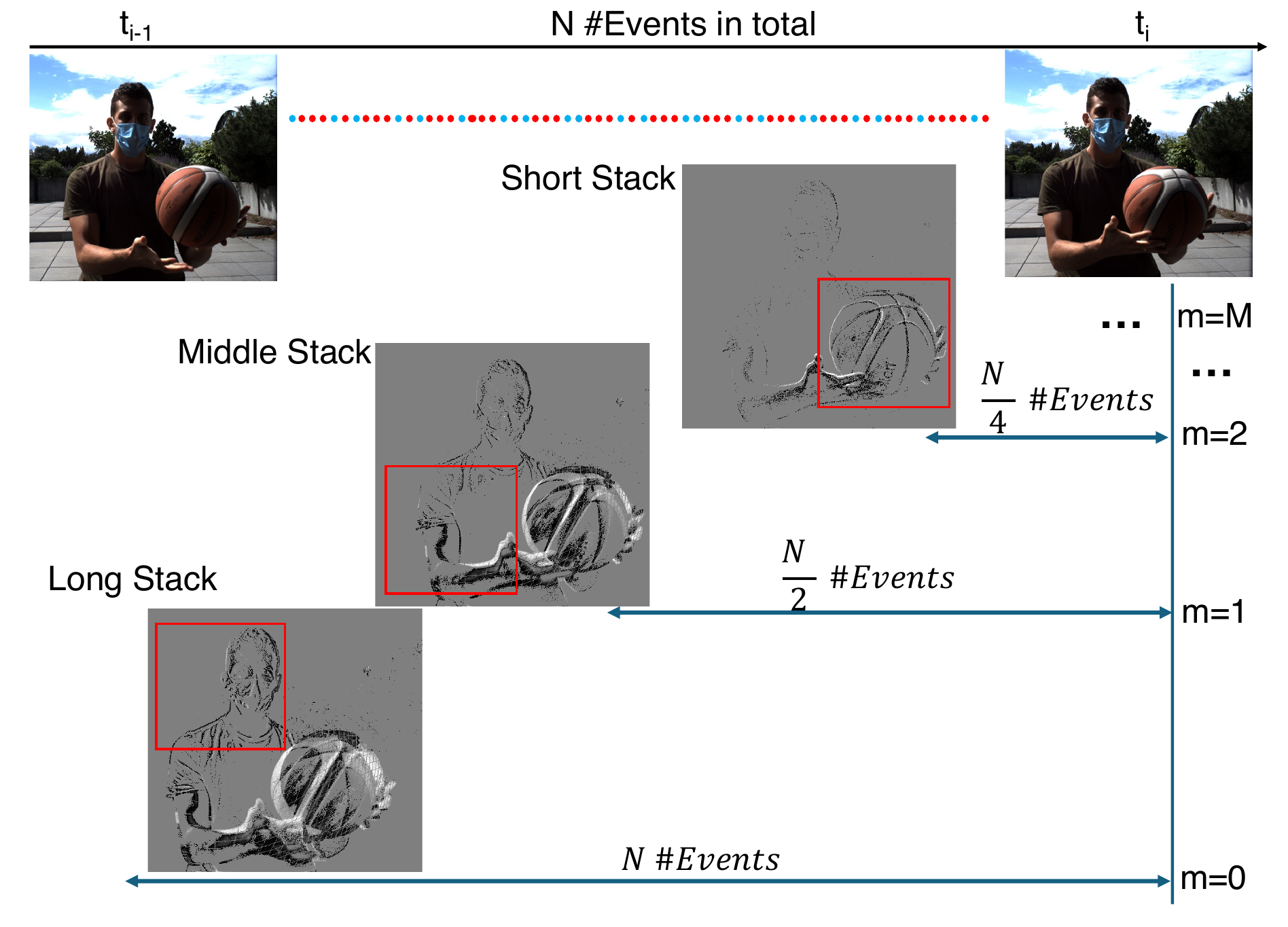}  

    \caption{Our multi-stack event representation is illustrated as follows. The stack begins from the target frame at $t_{i}$ and expands backward in time to the previous frame $t_{i-1}$. Within each stack, the number of events accumulated from $t_{i}$ is halved from the previous stack. In the long stack ($m=0$), the slowest-moving objects, such as around the human head, appear sharp; in the middle stack ($m=1$), slower-moving objects, like the human arm, are clear; and in the short stack ($m=2$), the fastest-moving objects, such as the ball, are sharp. This approach ensures that the event data provides adequate control information for generating frame $t_{i}$.}

    \label{fig:event_rep}
    \vspace{-5pt}
\end{figure}

To introduce events as the control condition for the video generation process, we incorporate two main components: 1) Event Representation and 2) Event Encoder. For event representation, we adopt a multi-stack approach, inspired by \cite{nam2022stereo, lu2024hr}, which captures both fast- and slow-moving objects within a multi-channel, frame-like format. An example of this multi-stack event representation is illustrated in Figure \ref{fig:event_rep}. Based on this representation, our event encoder consists of a series of convolutional layers with stride to downsample the input event representation into the event latent $E_{cond}$, matching the dimension of the image latent $I_{cond}$.

\begin{figure}[!t]
    \centering
    \includegraphics[width=.999\linewidth]{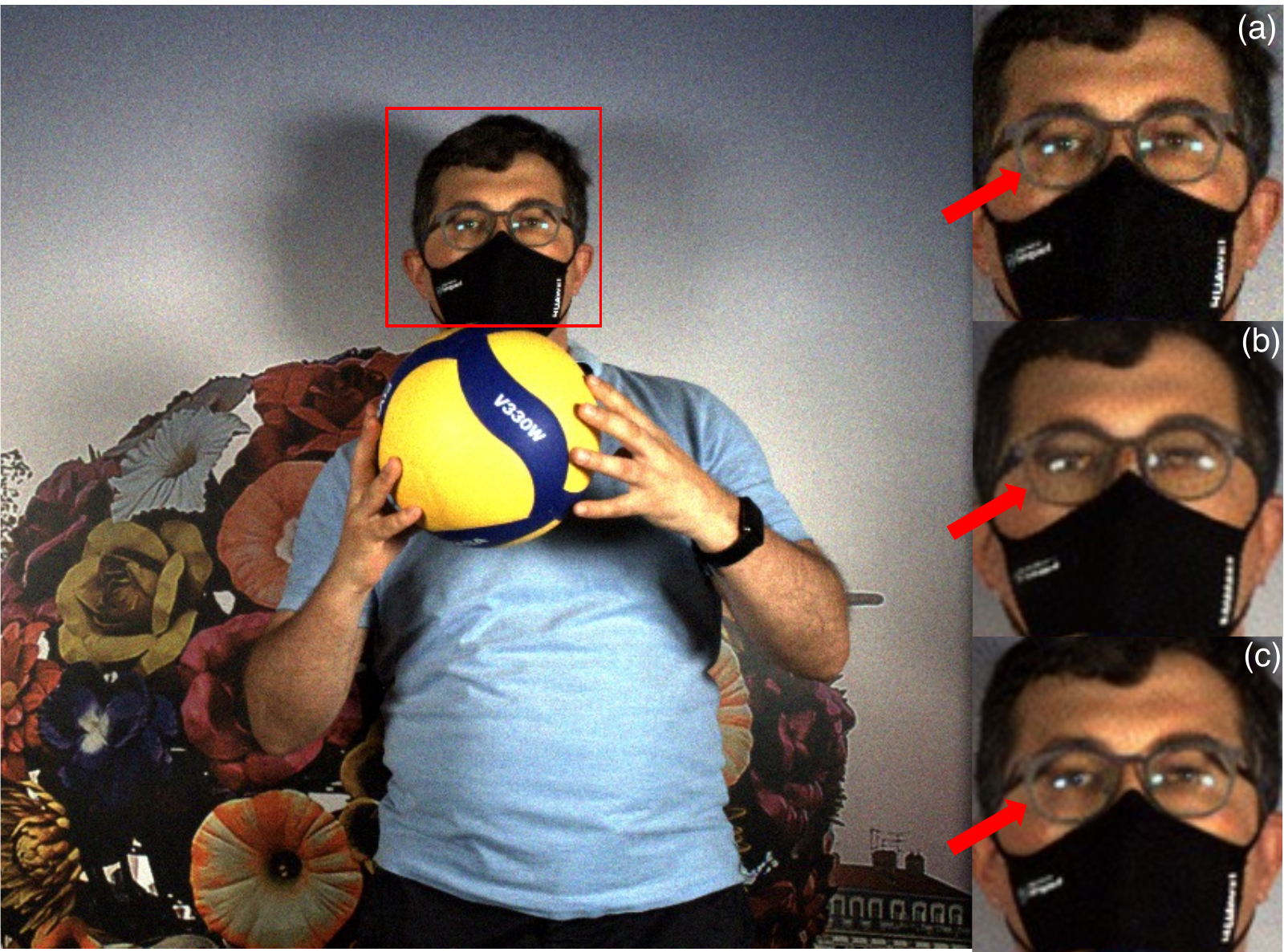}  

    \caption{The VAE encoding/decoding loss of small details in the original input image. In (a), we show the original image, sized 970 x 625. In (b), we pad the image to the nearest multiple of 8, then encode and decode it back. After decoding, the PSNR drops to 21.50, with noticeable detail loss in the zoomed-in view. In (c), we pad to the nearest multiple of 8, upsample to twice the original width and height, then encode and decode. After decoding, the PSNR increases to around 24.92, with no significant loss of details in the zoomed-in view.}

    \label{fig:upsample_tile}
    \vspace{-5pt}
\end{figure}

\subsection{Per-tile Denoising and Fusion}
\label{sec:overlap_tile}
Due to the substantial memory demands of training and generating high-resolution video, most existing video diffusion models, such as Stable Video Diffusion (SVD) \cite{blattmann2023stable}, are Latent Diffusion Models, meaning they operate in the latent space with downsampled spatial resolution rather than in the original pixel space. As illustrated in Figure \ref{fig:EVDS_train}, for an input RGB image/video of shape [$B$, $F$, 3, $H$, $W$], where $B$ is the batch size, $F$ is the number of frames, $W$ is the width, and $H$ is the height, a latent encoder ($\mathcal{E}$) and decoder ($\mathcal{D}$), typically based on a Variational Autoencoder (VAE), convert the input into image latents of shape [$B$, $F$, $C_{latent}$, $\frac{H}{d}$, $\frac{W}{d}$] and decode it back. 
% Here, $d$ is the downsample factor, set to 8 in SVD, and $C_{latent}$ is the number of latent channels, set to 4 in SVD. 

The Latent VAE encoder/decoder approach is effective for video generation, where realism is less critical, as the task often involves stylized animation rather than high-detail photorealism. However, for EVFI, minimizing the loss from downsampled encoding/decoding is essential, as our goal is to recover realistic missing frames between keyframes. A key observation is that small details can be lost after VAE encoding/decoding. Upsampling the input mitigates this loss, as shown in Figure \ref{fig:upsample_tile}, where increasing input upsampling effectively reduces encoding/decoding loss on fine details.

Additionally, as illustrated in Figure \ref{fig:EVDS_train}, spatial downsampling of the input image to create image latents can also reduce event control accuracy, since the trainable copy’s input consists of combined image latents $I_{cond}$ and event latents $E_{cond}$ of the same dimensions. However, upsampling input videos substantially increases computational costs for both training and testing. To address this, we divide the upsampled input video into overlapping cropped tiles of fixed width and height, applying EVDS to each tile independently. We then fuse the denoised latents of tiles to achieve consistent latents for each image and video, preserving shared context across overlapping regions

\begin{figure}[!t]
    \centering
    \includegraphics[width=.999\linewidth]{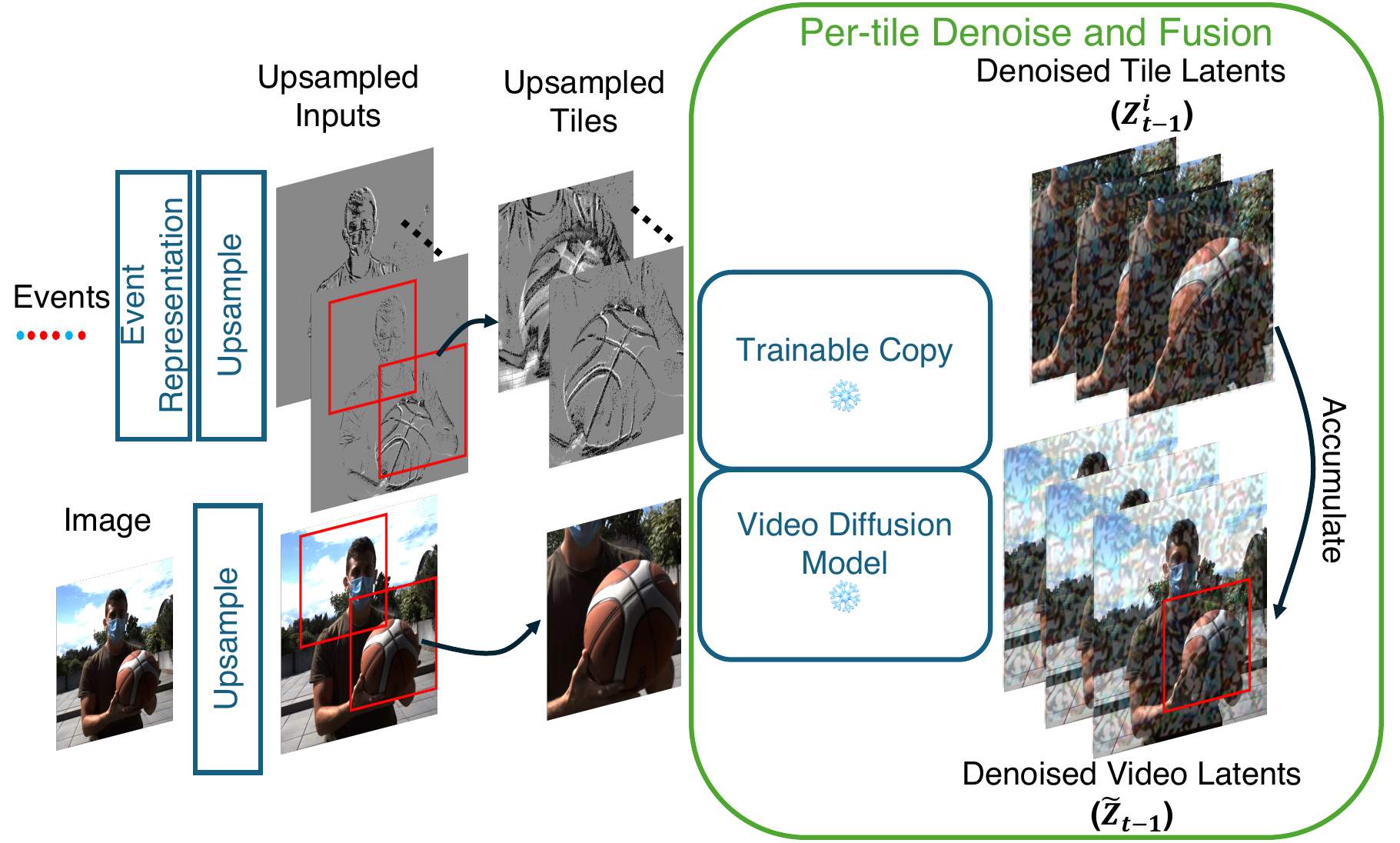}  

    \caption{The Per-tile Denoising and Fusion process is a test-time optimization applied during inference to enhance the fidelity of video generation appearance and improve event-based motion control accuracy. During each denoising step, our model operates on upsampled tiles of the input image and event representations to predict denoised latents for each tile $i$ at a denoising time $t$ ($Z_{t-1}^{i}$). These denoised tile latents are then accumulated to obtain the predicted denoised latents for the entire video (${\Tilde{Z}}_{t-1}$) at that denoising step $t$. }

    \label{fig:multidiffusion}
    \vspace{-5pt}
\end{figure}

Our Per-tile Denoising and Fusion process is illustrated in Figure \ref{fig:multidiffusion}. We employ tiled diffusion~\cite{bar2023multidiffusion} by first dividing the latent canvas into $n$ overlapping tiles. For the denoised tile latents at tile $i$ at denoising step $t$, we use an accumulation process to fuse them into the denoised latent for the entire upsampled video. The accumulation process takes in $n$ tile latents and outputs the fused latent:

\begin{equation}
    \Tilde{Z}_{t- 1} = \sum_{i=1}^{n} \frac{W_i}{\sum_{j = 1}^{n} W_j} \otimes Z^{i}_{t - 1},
\end{equation}
where $\Tilde{Z}_{t-1} \in \mathbb{R}^{F \times H \times W \times d}$, with $F$ representing the number of frames, $H$ and $W$ the height and width of the upsampled latent canvas, and $d$ the latent dimension. $Z^{i}_{t-1}$ contains the denoised latents of the $i$-th grid tile, $Z^{i}_{t-1} \in \mathbb{R}^{F \times h \times w \times d}$, where $h$ and $w$ denote the height and width of each latent tile. $W_i \in \mathbb{R}^{F \times h \times w}$ represents pixel-wise weights for the $i$-th tile. In our implementation, we assign equal weights, averaging the denoised latents across overlapping tiles.

\begin{table*}[t!]
    \centering
    \resizebox{1.0\linewidth}{!}{
    \begin{tabular}{@{}ccccccccccc@{}}
    \midrule
    \midrule
    \multirow{2}{*}{\textbf{Method}} & \multirow{2}{*}{\textbf{Category}}  \  &  \multicolumn{3}{|c|}{\textbf{BS-ERGB (3 skips)}}  \  &  \multicolumn{3}{c|}{\textbf{HQF (3 skips)}}  \  &  \multicolumn{3}{c|}{\textbf{Clear-Motion (15 skips)}} \\ 
    \cmidrule(r){3-11}
     & &  PSNR $\uparrow$ & SSIM $\uparrow$  & LPIPS $\downarrow$  & PSNR $\uparrow$  & SSIM $\uparrow$  & LPIPS $\downarrow$ & PSNR $\uparrow$  & SSIM $\uparrow$  & LPIPS $\downarrow$  \\ 
        \midrule
        Time Reversal \cite{feng2025explorative} (ECCV’24) & Video Diffusion \   &  17.86
                & 0.59   & 0.27 \    &  17.92
                & 0.53   & 0.18 \    &  13.71
                & 0.70  & 0.44 \\
        DynamiCrafter \cite{xing2024dynamicrafter} (ECCV’24)  & Video Diffusion \   & 15.47 
                &  0.54   & 0.40 \   & 15.99
                &  0.56   & 0.35 \  & 11.42
                &  0.62  & 0.52   \\

        \midrule
        RIFE \cite{huang2022real} (ECCV’22) & Frame-only VFI \    & 23.30
                & 0.83   &   \textcolor{blue}{0.09} \  & 25.34
                & 0.76   & \textcolor{blue}{0.05}  \    & 16.93
                & 0.76   & 0.35  \\  
        
        PerVFI \cite{Wu_2024_CVPR} (CVPR'24) & Frame-only VFI  &  23.53
        & 0.79   & 0.11 \    &  27.13
        & 0.81   & \textcolor{red}{0.04} \    &  16.74
        & 0.75  & \textcolor{blue}{0.30} \\
        InterpAny-Clearer \cite{zhong2024clearer}  (ECCV'24) & Frame-only VFI  \    & 25.51
                & 0.85   &   0.14 \  & 27.93
                & 0.83   & 0.06  \    & 17.21
                & 0.77   & 0.32  \\  
        
        EMA-VFI \cite{zhang2023extracting} (CVPR’23) & Frame-only VFI \   & 26.01
                &  \textcolor{blue}{0.86}   & 0.18 \   & 28.31
                &  \textcolor{blue}{0.84}   & 0.10 \  & 17.79
                &  0.78  & 0.44   \\
       GIMM-VFI-R-P \cite{guo2024generalizable} (NeurIPS’24) & Frame-only VFI \   & 25.56
                &  0.84   & \textcolor{red}{0.08} \   & 27.96
                &  0.83   & \textcolor{red}{0.04} \  & 17.77
                &  0.77  & \textcolor{red}{0.26}   \\
        \midrule
        CBMNet-Large \cite{Kim_2023_CVPR} (CVPR’23) & EVFI \   & \textcolor{blue}{26.24} 
                &  0.79   & 0.22 \   & \textcolor{blue}{28.73} 
                &  \textcolor{blue}{0.84}   & 0.09 \  & \textcolor{blue}{22.26} 
                &  \textcolor{blue}{0.86}  & 0.39   \\ 
         RE-VDM (Ours) & EVFI + Video Diffusion   &  \textcolor{red}{27.74}  & \textcolor{red}{0.88} & 0.12 \ & \textcolor{red}{29.04}  & \textcolor{red}{0.89} & 0.06  \ & \textcolor{red}{22.94}  & \textcolor{red}{0.88} & 0.37 \\
        \midrule
        \midrule
    \end{tabular}
    }
    \captionsetup{font=small}
    \caption{We compare our method on real-world EVFI datasets against representative and SOTA methods, including the pre-trained video diffusion model (Video Diffusion), frame-only methods (Frame-only VFI), and event-based video frame interpolation method (EVFI), where \textcolor{red}{red} indicates the best metric and \textcolor{blue}{blue} indicates the second-best. For CBMNet-Large, we use publicly available model checkpoints trained on the same BS-ERGB dataset as our approach. For the BS-ERGB and HQF datasets, we use a skip number of 3 as in the EVFI papers, we evaluate all interpolated frames. For our Clear-Motion test sequences, we use a skip number of 15 for evaluation on large motion scenarios, we evaluate the 4th, 8th, and 12th interpolated frames. To control for VAE encoding/decoding loss, as discussed in Section \ref{sec:eval_strategy}, we apply VAE encoding/decoding to all model outputs.}
    \label{tab:compareResults}
\end{table*}

\subsection{From Video Generation to Frame Interpolation}
\label{sec:interpolation}
Our approach thus far is designed to generate realistic video controlled by events from the start frame, suited for event-based video generation (extrapolation). However, the EVFI task requires frame interpolation, which involves utilizing information from both the start and end frames.

The most straightforward approach to achieve frame interpolation would be to fine-tune base video diffusion models to accept an additional channel for the second image. However, as previously discussed, fine-tuning pre-trained video diffusion foundation models on the EVFI dataset risks catastrophic forgetting of the learned data priors, undermining the purpose of adaptation.

To address this, we opted for a test-time optimization approach that avoids further training, enabling our video generation model to perform EVFI. As shown in Figure \ref{fig:pipe_overview}, inspired by \cite{feng2025explorative}, our key idea is to run the EVDS at each denoising step from both the start frame $I_{s}$ and the end frame $I_{e}$, producing two sets of predicted denoised video latents. We then apply a Two-side Fusion of these denoised latents to obtain consistent results. This approach allows us to achieve frame interpolation, effectively leveraging information from both frames. The details of the Two-Side Fusion algorithm are as follows: 
For a denoising time $t$, the forward denoised video latents from the start frame $I_{s}$ and forward-time events $E_{s}$ are $Z^{s}_{t - 1} = EVDS(I_{s}, E_{s})$, and the backward denoised video latents from the end frame $I_{e}$ and backward-time events are $Z^{e}_{t - 1} = EVDS(I_{e}, E_{e})$, with $Z^{s}_{t - 1}, Z^{e}_{t - 1} \in \mathbb{R}^{F \times H \times W \times d}$. To make the final denoised video latent $\Tilde{Z}_{t-1}$ consistent with information from both the start and end frames, we assign linearly increasing weights, $W_{f} \in \mathbb{R}^{F}$, where $W_{0} = 0$ and $W_{1} = 1$. This allows us to fuse the final video latents $\Tilde{Z}_{t-1}$ as follows:
\begin{equation}
    \Tilde{Z}_{t- 1} =  W_{f} \otimes Z^{s}_{t - 1} + (1 -  W_{f}) \otimes flip(Z^{e}_{t - 1})
\end{equation}
where $flip$ denotes reversing the temporal order of the frames in $Z^{e}_{t-1}$.
The intuition behind this approach is that frames near the start frame rely more heavily on information from the start frame and forward-time events, while frames near the end rely more on the end frame and backward-time events, in the middle, they balance information from both.

% This forward-backward consistent denoising step of a frame latent in our diffusion process generalizes the traditional EVFI methods \cite{tulyakov2021time, tulyakov2022time, Kim_2023_CVPR} by leveraging forward-time event voxels from the start frame to $t_{interp}$ and backward-time event voxels from the end frame to interpolate the frame at $t_{interp}$ to ensure consistent frame predictions from both directions.

% \begin{itemize}
%     \item Data-Efficient Adaption of Pre-trained Video Diffusion Models to Event-based video extrapolation. (ControlNet)
%     \item Event Representation to convert event stream into multi-channel image like representation for slow and fast moving objects
%     \item  Spatial Resolution Issue of Latnet Diffusion Model (VAE) Downsampling in image and contorl condition, and multi-diffusion to solve it. 
%     \item  Converting Extroplation into interpolation by enforcing test-time optimization of forward-backward consistency for generantion from two sides.
% \end{itemize}

% You must include your signed IEEE copyright release form when you submit your finished paper.
% We MUST have this form before your paper can be published in the proceedings.

% Please direct any questions to the production editor in charge of these proceedings at the IEEE Computer Society Press:
% \url{https://www.computer.org/about/contact}.
\section{Experimental Results}

\subsection{Datasets and Implementation Details}
To robustly evaluate the generalized frame interpolation performance of our approach, we select several real-world event camera datasets. The BS-ERGB dataset \cite{tulyakov2021time}, the most cited for EVFI, provides aligned RGB and event video with large motion between frames, mainly featuring static cameras capturing dynamic objects. The HQF dataset \cite{stoffregen2020reducing} includes grayscale video aligned with events, with more sequences involving a moving event camera. In addition to existing datasets, we include our self-collected Clear-Motion test sequences, capturing both object and camera motion. Detailed information is provided in the Supplementary Material. Clear-Motion is a self-collected dataset used solely for testing, eliminating the possibility of data leakage. This makes it ideal for robustly evaluating the zero-shot generalization performance of all models. The pre-trained video diffusion model we used is Stable Video Diffusion \cite{blattmann2023stable} for 14-frame image-to-video generation. We provide detailed training information in the supplementary material.
% We trained our model with an effective batch size of 64, using a batch size of 4 per GPU and a gradient accumulation factor of 16. Training was conducted solely on the BS-ERGB dataset, and the model was tested on other unseen datasets without fine-tuning. All training was performed on 4 NVIDIA RTX A6000 GPUs, each with 50GB of memory.

\subsection{Evaluation Strategy}
\label{sec:eval_strategy}
To evaluate generalized frame interpolation performance, we compare our approach with representative methods from three categories: frame-only interpolation, pre-trained video diffusion models, and event-based interpolation. 
We use publicly available CBMNet checkpoints trained on the BS-ERGB dataset for evaluation. 
For Per-tile Denoising and Fusion in our method, which uses upsampled tiles to reduce detail loss and enhance event control accuracy, we apply an upsample factor of 2 for the BS-ERGB and Clear-Motion datasets and 3 for the HQF dataset.

For evaluation metrics, we use average frame PSNR \cite{gonzalez2009digital}, SSIM \cite{wang2004image}, and LPIPS \cite{zhang2018unreasonable} for the interpolated frames. Since SVD is a latent diffusion model that uses a VAE to encode and decode between latent and pixel space, we apply VAE encoding and decoding to all model outputs. This eliminates the impact of discrepancies between the VAE-decoded frame distribution and the ground truth frame distribution, accounting for differences in small details and tone mapping.

% \subsection{Implementation Details}
% \begin{itemize}
%     \item Model: SVD- Latent Diffusion Model
%     \item All real-world datasets: BSERGB (\checkmark), HQF (\checkmark), THU-HSEVI. Lastly our real capture data for quantitative/qualitative evaluation.
%     \item Metrics: per-frame quality: PSNR, SSIM, LPIPS, consistency: Warp-loss, X-t Plot (Visualization). *To take into account of the VAE encode/decode loss of per-frame quality (tonemapping, smoothing), we apply VAE encoding/decoding to all models' outputs
% \end{itemize}

\subsection{Evaluation on Video Frame Interpolation}

\begin{figure*}[t]
    \centering
    \includegraphics[width=0.95\linewidth]{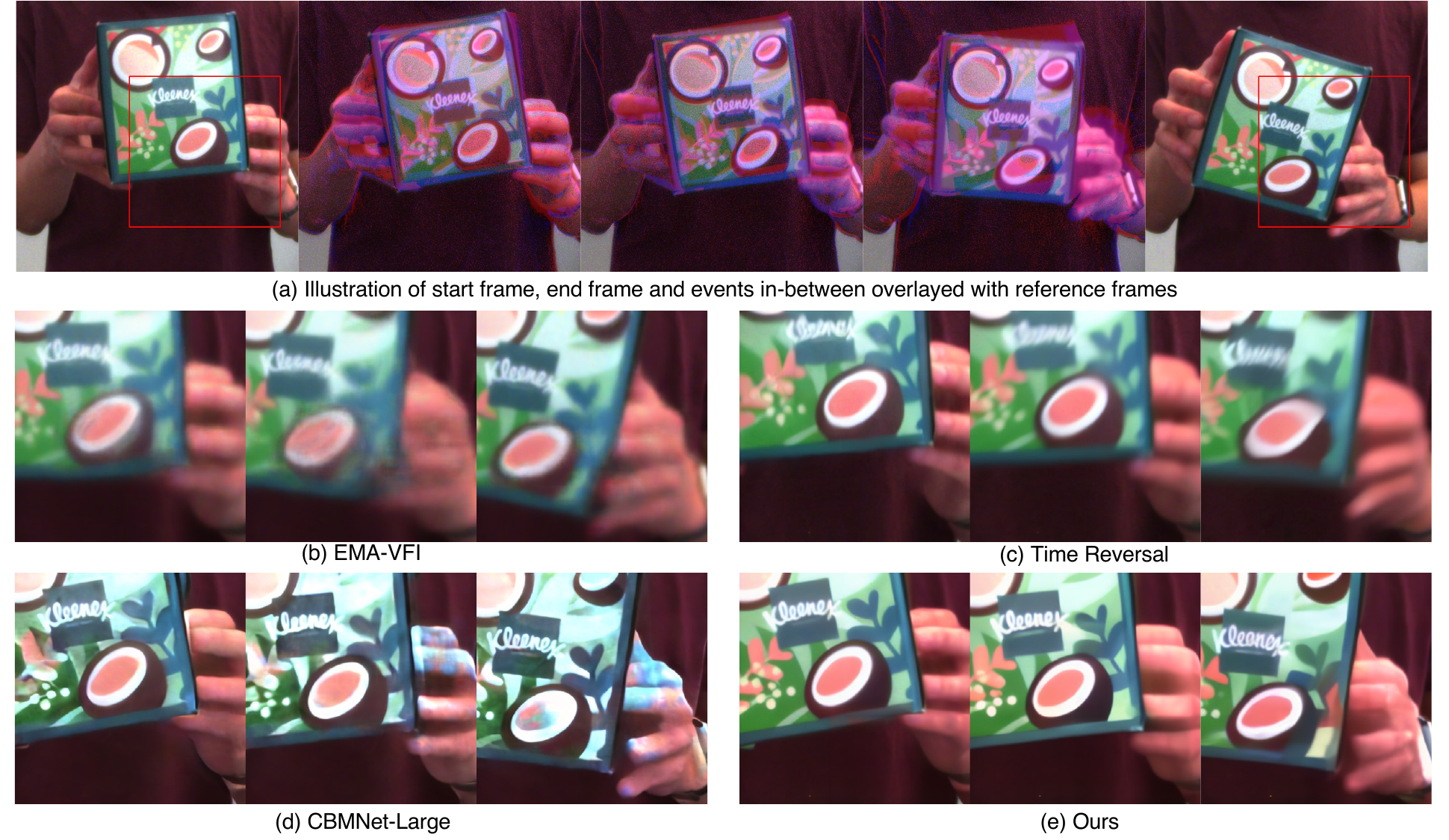}

    \caption{The qualitative comparison on the Clear-Motion sequence Texture\_Box, featuring translation and rotation of a textured box, demonstrates that our method achieves the most consistent and accurate frame reconstruction. Other methods experience significant detail loss, artifacts, and difficulty maintaining regular object shapes, particularly in the characters and the bottom-right coconut. }

    \label{fig:real_box}
    \vspace{-7pt}
\end{figure*}

\begin{figure*}[t]
    \centering
    \includegraphics[width=.95\linewidth]{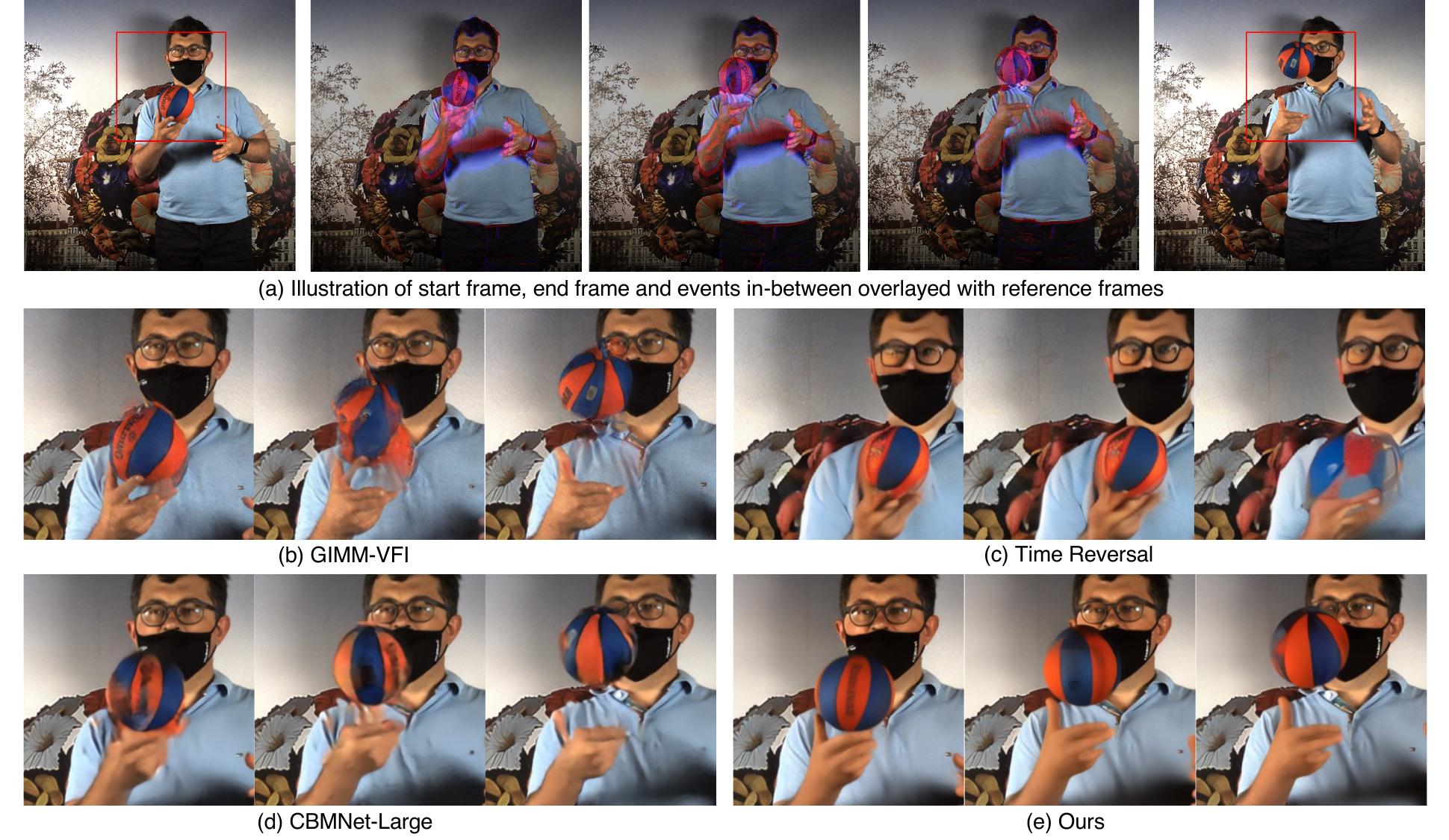}  

    \caption{The qualitative comparison on the BS-ERGB test sequences, featuring fast motion of a hand and a rising, spinning ball, shows that our method is the only approach capable of maintaining the regular texture shapes on both the ball and hand in this challenging scenario. }

    \label{fig:real_checker}
    \vspace{-5pt}
\end{figure*}

In this section, we present both quantitative and qualitative results demonstrating the effectiveness of our method compared to representative video frame interpolation approaches, as shown in Table \ref{tab:compareResults}. Our method outperforms all recent SOTA VFI baselines in terms of PSNR and SSIM. Qualitative comparisons in Figures \ref{fig:real_box} and \ref{fig:real_checker} further show that recent frame-only VFI methods struggle with large and complex motion due to the ill-posed nature of the task. This limitation also affects pre-trained video diffusion models, which often exhibit a bias toward animation-style generation and suffer from spatial-temporal detail loss.

In contrast, the SOTA event-based interpolation method, CBMNet-Large, can handle large motions by leveraging cues from events, similar to our approach. However, its performance on both the test set from the training dataset and unseen datasets is consistently lower than ours across all metrics. This difference is due to two factors: 1) as discussed in Section \ref{sec:intro} and Table \ref{tab:compare_fields}, the pre-trained video diffusion foundation model in our approach offers a superior data prior from its extensive training on large video generation datasets, and 2) our model's backbone, a highly representative pre-trained video diffusion model, provides advantages over custom models typically used in EVFI.

The effectiveness of our approach on real-world, unseen data for video frame interpolation is clearly demonstrated in the qualitative comparisons in Figure \ref{fig:real_box}. Due to the ill-posed nature of interpolating frames with large motion, frame-only methods, despite often achieving better LPIPS scores, struggle to accurately reconstruct object locations and appearance.
While CBMNet-Large uses event data to guide interpolation and achieves similar object and texture placements, it lacks generalization on unseen object and camera motion compared to our method, as shown in Figure  \ref{fig:real_box}, despite being trained on the same dataset. Additionally, CBMNet-Large suffers from frame-to-frame inconsistencies due to its frame-by-frame inference design, evident in Figure \ref{fig:real_checker}. CBMNet-Large produces frames with inconsistent textures and shapes on the ball and hand, whereas our method accurately reconstructs the challenging, regular shapes of both. This advantage is due to the strong data priors and foundational design of the pre-trained video diffusion model, allowing for consistent reconstruction across frames. These results support our unique contributions and motivation in adapting pre-trained video diffusion models for EVFI.
\subsection{Ablation Study}
In the Supplementary Material, we provide quantitative and qualitative ablation studies on two key components for adapting pre-trained VDMs to EVFI: 1) Comparing video generation and video interpolation using Two-Side Fusion. 2) Ablating the effect of upsampling in Per-tile Denoising and Fusion. Both quantitative and qualitative results demonstrate the effectiveness of the two main components in our approach for adapting pre-trained VDMs to the EVFI task.

% \begin{figure*}[t]
%     \centering
%     \includegraphics[width=1\linewidth]{sec/Figures/Supp/exp_Supp_1.pdf}  

%     \caption{The qualitative visualization of our experimental results on the Clear-Motion test sequences, featuring large camera motion, demonstrates that our method achieves the most consistent and accurate frame reconstruction. In contrast, other methods suffer from significant detail loss, artifacts, and challenges in preserving regular object shapes, especially in the edges and details of cabinets and small objects. }

%     \label{fig:real_box}
%     \vspace{-7pt}
% \end{figure*}

\section{Limitations}

Despite the promising results, adapting pre-trained video diffusion models to the EVFI task presents several limitations. First, the pre-trained VAE encoder and decoder, used for conversion between pixel space and latent space, have limited representational power, often resulting in color shifts and loss of small details. Second, the adapted video diffusion model tends to smooth out certain regions, likely due to its difficulty in preserving fine textures within the downsampled latent space or the need for a prohibitively high upsampling factor to prevent smoothing. Finally, the diffusion-based methods are more memory-intensive and time-consuming than other methods, we provide the detailed model run time, memory comparison in the Supplementary Material. These limitations stem from the current state of pre-trained video diffusion foundation models. However, as the field of video generation evolves rapidly, we anticipate that many of these issues could be mitigated or resolved with newer, more advanced pre-trained models in the near future.

\section{Conclusion}
In this work, we explore, for the first time, the adaptation of pre-trained video diffusion foundation models for the event-based video frame interpolation (EVFI) task, as a foundational step toward integrating event camera technology with the era of Generative AI. We present an approach to address the key challenges in adapting video diffusion models to EVFI.  We use data-efficient adaptation to bridge the gap in data quality and size between large-scale video generation datasets and smaller EVFI datasets, minimizing the risk of catastrophic forgetting. Our method also includes test-time optimization techniques: Per-tile Denoising and Fusion to reduce encoding/decoding losses between pixel and latent spaces, preserving high-fidelity appearance and accurate event-based motion control, and Two-side Fusion, which incorporates both start and end frame information at each denoising step for improved interpolation consistency. We also discuss current limitations in pre-trained video diffusion models for EVFI. Lastly, we demonstrate our model’s superior generalization and frame consistency on unseen real-world data, highlighting the motivation and potential of our approach.

\section{Acknowledgment}
We greatly acknowledge NSF’s support under awards OISE 2020624 and BCS 2318255

\newpage

{
    \small
    \bibliographystyle{ieeenat_fullname}
    \bibliography{main}

\begin{thebibliography}{55}
\providecommand{\natexlab}[1]{#1}
\providecommand{\url}[1]{\texttt{#1}}
\expandafter\ifx\csname urlstyle\endcsname\relax
  \providecommand{\doi}[1]{doi: #1}\else
  \providecommand{\doi}{doi: \begingroup \urlstyle{rm}\Url}\fi

\bibitem[Bao et~al.(2019)Bao, Lai, Ma, Zhang, Gao, and Yang]{bao2019depth}
Wenbo Bao, Wei-Sheng Lai, Chao Ma, Xiaoyun Zhang, Zhiyong Gao, and Ming-Hsuan Yang.
\newblock Depth-aware video frame interpolation.
\newblock In \emph{Proceedings of the IEEE/CVF conference on computer vision and pattern recognition}, pages 3703--3712, 2019.

\bibitem[Bar-Tal et~al.(2023)Bar-Tal, Yariv, Lipman, and Dekel]{bar2023multidiffusion}
Omer Bar-Tal, Lior Yariv, Yaron Lipman, and Tali Dekel.
\newblock Multidiffusion: Fusing diffusion paths for controlled image generation.
\newblock In \emph{ICML}. PMLR, 2023.

\bibitem[Bardow et~al.(2016)Bardow, Davison, and Leutenegger]{bardow2016simultaneous}
Patrick Bardow, Andrew~J Davison, and Stefan Leutenegger.
\newblock Simultaneous optical flow and intensity estimation from an event camera.
\newblock In \emph{Proceedings of the IEEE conference on computer vision and pattern recognition}, pages 884--892, 2016.

\bibitem[Barranco et~al.(2014)Barranco, Ferm{\"u}ller, and Aloimonos]{barranco2014contour}
Francisco Barranco, Cornelia Ferm{\"u}ller, and Yiannis Aloimonos.
\newblock Contour motion estimation for asynchronous event-driven cameras.
\newblock \emph{Proceedings of the IEEE}, 102\penalty0 (10):\penalty0 1537--1556, 2014.

\bibitem[Barranco et~al.(2018)Barranco, Ferm\"uller, and Ros]{barranco2018real}
Francisco Barranco, Cornelia Ferm\"uller, and Eduardo Ros.
\newblock Real-time clustering and multi-target tracking using event-based sensors.
\newblock In \emph{2018 IEEE/RSJ International Conference on Intelligent Robots and Systems (IROS)}, pages 5764--5769. IEEE, 2018.

\bibitem[Benosman et~al.(2012)Benosman, Ieng, Clercq, Bartolozzi, and Srinivasan]{benosman2012asynchronous}
Ryad Benosman, Sio-Hoi Ieng, Charles Clercq, Chiara Bartolozzi, and Mandyam Srinivasan.
\newblock Asynchronous frameless event-based optical flow.
\newblock \emph{Neural Networks}, 27:\penalty0 32--37, 2012.

\bibitem[Blattmann et~al.(2023{\natexlab{a}})Blattmann, Dockhorn, Kulal, Mendelevitch, Kilian, Lorenz, Levi, English, Voleti, Letts, et~al.]{blattmann2023stable}
Andreas Blattmann, Tim Dockhorn, Sumith Kulal, Daniel Mendelevitch, Maciej Kilian, Dominik Lorenz, Yam Levi, Zion English, Vikram Voleti, Adam Letts, et~al.
\newblock Stable video diffusion: Scaling latent video diffusion models to large datasets.
\newblock \emph{arXiv preprint arXiv:2311.15127}, 2023{\natexlab{a}}.

\bibitem[Blattmann et~al.(2023{\natexlab{b}})Blattmann, Rombach, Ling, Dockhorn, Kim, Fidler, and Kreis]{blattmann2023align}
Andreas Blattmann, Robin Rombach, Huan Ling, Tim Dockhorn, Seung~Wook Kim, Sanja Fidler, and Karsten Kreis.
\newblock Align your latents: High-resolution video synthesis with latent diffusion models.
\newblock In \emph{CVPR}, 2023{\natexlab{b}}.

\bibitem[Cadena et~al.(2023)Cadena, Qian, Wang, and Yang]{cadena2023sparse}
Pablo Rodrigo~Gantier Cadena, Yeqiang Qian, Chunxiang Wang, and Ming Yang.
\newblock Sparse-e2vid: A sparse convolutional model for event-based video reconstruction trained with real event noise.
\newblock In \emph{Proceedings of the IEEE/CVF Conference on Computer Vision and Pattern Recognition}, pages 4149--4157, 2023.

\bibitem[Chen et~al.(2024)Chen, Feng, Cai, Xie, Metzler, Fermuller, and Aloimonos]{chen2024timerewind}
Jingxi Chen, Brandon~Y Feng, Haoming Cai, Mingyang Xie, Christopher Metzler, Cornelia Fermuller, and Yiannis Aloimonos.
\newblock Timerewind: Rewinding time with image-and-events video diffusion.
\newblock \emph{arXiv preprint arXiv:2403.13800}, 2024.

\bibitem[Feng et~al.(2024)Feng, Ding, Xia, Niklaus, Abrevaya, Black, and Zhang]{feng2025explorative}
Haiwen Feng, Zheng Ding, Zhihao Xia, Simon Niklaus, Victoria Abrevaya, Michael~J Black, and Xuaner Zhang.
\newblock Explorative inbetweening of time and space.
\newblock In \emph{ECCV}, 2024.

\bibitem[Gallego et~al.(2020)Gallego, Delbr{\"u}ck, Orchard, Bartolozzi, Taba, Censi, Leutenegger, Davison, Conradt, Daniilidis, et~al.]{gallego2020event}
Guillermo Gallego, Tobi Delbr{\"u}ck, Garrick Orchard, Chiara Bartolozzi, Brian Taba, Andrea Censi, Stefan Leutenegger, Andrew~J Davison, J{\"o}rg Conradt, Kostas Daniilidis, et~al.
\newblock Event-based vision: A survey.
\newblock \emph{IEEE transactions on pattern analysis and machine intelligence}, 44\penalty0 (1):\penalty0 154--180, 2020.

\bibitem[Gonzalez(2009)]{gonzalez2009digital}
Rafael~C Gonzalez.
\newblock \emph{Digital image processing}.
\newblock Pearson education india, 2009.

\bibitem[Guo et~al.(2024)Guo, Li, and Loy]{guo2024generalizable}
Zujin Guo, Wei Li, and Chen~Change Loy.
\newblock Generalizable implicit motion modeling for video frame interpolation.
\newblock \emph{Advances in Neural Information Processing Systems}, 37:\penalty0 63747--63770, 2024.

\bibitem[He et~al.(2024)He, Wang, Zhou, Chen, Singh, Li, Gao, Shen, Wang, Cao, et~al.]{he2024microsaccade}
Botao He, Ze Wang, Yuan Zhou, Jingxi Chen, Chahat~Deep Singh, Haojia Li, Yuman Gao, Shaojie Shen, Kaiwei Wang, Yanjun Cao, et~al.
\newblock Microsaccade-inspired event camera for robotics.
\newblock \emph{Science Robotics}, 9\penalty0 (90):\penalty0 eadj8124, 2024.

\bibitem[Ho et~al.(2020)Ho, Jain, and Abbeel]{ho2020denoising}
Jonathan Ho, Ajay Jain, and Pieter Abbeel.
\newblock Denoising diffusion probabilistic models.
\newblock In \emph{NeurIPS}, 2020.

\bibitem[Ho et~al.(2022)Ho, Salimans, Gritsenko, Chan, Norouzi, and Fleet]{ho2022video}
Jonathan Ho, Tim Salimans, Alexey Gritsenko, William Chan, Mohammad Norouzi, and David~J Fleet.
\newblock Video diffusion models.
\newblock 2022.

\bibitem[Huang et~al.(2022)Huang, Zhang, Heng, Shi, and Zhou]{huang2022real}
Zhewei Huang, Tianyuan Zhang, Wen Heng, Boxin Shi, and Shuchang Zhou.
\newblock Real-time intermediate flow estimation for video frame interpolation.
\newblock In \emph{European Conference on Computer Vision}, pages 624--642. Springer, 2022.

\bibitem[Jia et~al.(2024)Jia, Hoyer, Huang, Wang, Van~Gool, Schindler, and Obukhov]{jia2025dginstyle}
Yuru Jia, Lukas Hoyer, Shengyu Huang, Tianfu Wang, Luc Van~Gool, Konrad Schindler, and Anton Obukhov.
\newblock Dginstyle: Domain-generalizable semantic segmentation with image diffusion models and stylized semantic control.
\newblock In \emph{ECCV}, 2024.

\bibitem[Jiang et~al.(2018)Jiang, Sun, Jampani, Yang, Learned-Miller, and Kautz]{jiang2018super}
Huaizu Jiang, Deqing Sun, Varun Jampani, Ming-Hsuan Yang, Erik Learned-Miller, and Jan Kautz.
\newblock Super slomo: High quality estimation of multiple intermediate frames for video interpolation.
\newblock In \emph{Proceedings of the IEEE conference on computer vision and pattern recognition}, pages 9000--9008, 2018.

\bibitem[Kim et~al.(2023)Kim, Chae, Jang, and Yoon]{Kim_2023_CVPR}
Taewoo Kim, Yujeong Chae, Hyun-Kurl Jang, and Kuk-Jin Yoon.
\newblock Event-based video frame interpolation with cross-modal asymmetric bidirectional motion fields.
\newblock In \emph{Proceedings of the IEEE/CVF Conference on Computer Vision and Pattern Recognition (CVPR)}, pages 18032--18042, 2023.

\bibitem[Kong et~al.(2022)Kong, Jiang, Luo, Chu, Huang, Tai, Wang, and Yang]{kong2022ifrnet}
Lingtong Kong, Boyuan Jiang, Donghao Luo, Wenqing Chu, Xiaoming Huang, Ying Tai, Chengjie Wang, and Jie Yang.
\newblock Ifrnet: Intermediate feature refine network for efficient frame interpolation.
\newblock In \emph{Proceedings of the IEEE/CVF Conference on Computer Vision and Pattern Recognition}, pages 1969--1978, 2022.

\bibitem[Liu et~al.(2024)Liu, Zhang, Li, Yan, Gao, Chen, Yuan, Huang, Sun, Gao, et~al.]{liu2024sora}
Yixin Liu, Kai Zhang, Yuan Li, Zhiling Yan, Chujie Gao, Ruoxi Chen, Zhengqing Yuan, Yue Huang, Hanchi Sun, Jianfeng Gao, et~al.
\newblock Sora: A review on background, technology, limitations, and opportunities of large vision models.
\newblock \emph{arXiv preprint arXiv:2402.17177}, 2024.

\bibitem[Loshchilov et~al.(2017)Loshchilov, Hutter, et~al.]{loshchilov2017fixing}
Ilya Loshchilov, Frank Hutter, et~al.
\newblock Fixing weight decay regularization in adam.
\newblock \emph{arXiv preprint arXiv:1711.05101}, 5, 2017.

\bibitem[Lu et~al.(2022)Lu, Wu, Lin, Lu, and Jia]{lu2022video}
Liying Lu, Ruizheng Wu, Huaijia Lin, Jiangbo Lu, and Jiaya Jia.
\newblock Video frame interpolation with transformer.
\newblock In \emph{Proceedings of the IEEE/CVF Conference on Computer Vision and Pattern Recognition}, pages 3532--3542, 2022.

\bibitem[Lu et~al.(2024)Lu, Wang, Wang, and Xiong]{lu2024hr}
Yunfan Lu, Zipeng Wang, Yusheng Wang, and Hui Xiong.
\newblock Hr-inr: continuous space-time video super-resolution via event camera.
\newblock \emph{arXiv preprint arXiv:2405.13389}, 2024.

\bibitem[Lugmayr et~al.(2022)Lugmayr, Danelljan, Romero, Yu, Timofte, and Van~Gool]{lugmayr2022repaint}
Andreas Lugmayr, Martin Danelljan, Andres Romero, Fisher Yu, Radu Timofte, and Luc Van~Gool.
\newblock Repaint: Inpainting using denoising diffusion probabilistic models.
\newblock In \emph{CVPR}, pages 11461--11471, 2022.

\bibitem[Ma et~al.(2024)Ma, Guo, Chen, Xue, and Gu]{ma2025timelens}
Yongrui Ma, Shi Guo, Yutian Chen, Tianfan Xue, and Jinwei Gu.
\newblock Timelens-xl: Real-time event-based video frame interpolation with large motion.
\newblock In \emph{European Conference on Computer Vision}, pages 178--194. Springer, 2024.

\bibitem[Mitrokhin et~al.(2018)Mitrokhin, Ferm{\"u}ller, Parameshwara, and Aloimonos]{mitrokhin2018event}
Anton Mitrokhin, Cornelia Ferm{\"u}ller, Chethan Parameshwara, and Yiannis Aloimonos.
\newblock Event-based moving object detection and tracking.
\newblock In \emph{2018 IEEE/RSJ International Conference on Intelligent Robots and Systems (IROS)}, pages 1--9. IEEE, 2018.

\bibitem[Nam et~al.(2022)Nam, Mostafavi, Yoon, and Choi]{nam2022stereo}
Yeongwoo Nam, Mohammad Mostafavi, Kuk-Jin Yoon, and Jonghyun Choi.
\newblock Stereo depth from events cameras: Concentrate and focus on the future.
\newblock In \emph{Proceedings of the IEEE/CVF Conference on Computer Vision and Patter Recognition}, 2022.

\bibitem[Park et~al.(2021)Park, Lee, and Kim]{park2021asymmetric}
Junheum Park, Chul Lee, and Chang-Su Kim.
\newblock Asymmetric bilateral motion estimation for video frame interpolation.
\newblock In \emph{Proceedings of the IEEE/CVF international conference on computer vision}, pages 14539--14548, 2021.

\bibitem[Polyak et~al.(2024)Polyak, Zohar, Brown, Tjandra, Sinha, Lee, Vyas, Shi, Ma, Chuang, et~al.]{polyak2024movie}
Adam Polyak, Amit Zohar, Andrew Brown, Andros Tjandra, Animesh Sinha, Ann Lee, Apoorv Vyas, Bowen Shi, Chih-Yao Ma, Ching-Yao Chuang, et~al.
\newblock Movie gen: A cast of media foundation models.
\newblock \emph{arXiv preprint arXiv:2410.13720}, 2024.

\bibitem[Rebecq et~al.(2019)Rebecq, Ranftl, Koltun, and Scaramuzza]{rebecq2019high}
Henri Rebecq, Ren{\'e} Ranftl, Vladlen Koltun, and Davide Scaramuzza.
\newblock High speed and high dynamic range video with an event camera.
\newblock \emph{IEEE transactions on pattern analysis and machine intelligence}, 43\penalty0 (6):\penalty0 1964--1980, 2019.

\bibitem[Rombach et~al.(2022)Rombach, Blattmann, Lorenz, Esser, and Ommer]{rombach2022high}
Robin Rombach, Andreas Blattmann, Dominik Lorenz, Patrick Esser, and Bj{\"o}rn Ommer.
\newblock High-resolution image synthesis with latent diffusion models.
\newblock In \emph{CVPR}, 2022.

\bibitem[Ruiz et~al.(2023)Ruiz, Li, Jampani, Pritch, Rubinstein, and Aberman]{ruiz2023dreambooth}
Nataniel Ruiz, Yuanzhen Li, Varun Jampani, Yael Pritch, Michael Rubinstein, and Kfir Aberman.
\newblock Dreambooth: Fine tuning text-to-image diffusion models for subject-driven generation.
\newblock In \emph{Proceedings of the IEEE/CVF conference on computer vision and pattern recognition}, pages 22500--22510, 2023.

\bibitem[Schuhmann et~al.(2021)Schuhmann, Vencu, Beaumont, Kaczmarczyk, Mullis, Katta, Coombes, Jitsev, and Komatsuzaki]{schuhmann2021laion}
Christoph Schuhmann, Richard Vencu, Romain Beaumont, Robert Kaczmarczyk, Clayton Mullis, Aarush Katta, Theo Coombes, Jenia Jitsev, and Aran Komatsuzaki.
\newblock Laion-400m: Open dataset of clip-filtered 400 million image-text pairs.
\newblock \emph{arXiv preprint arXiv:2111.02114}, 2021.

\bibitem[Shah et~al.(2024)Shah, Chan, Cai, Chen, Kulshrestha, Singh, Aloimonos, and Metzler]{shah2024codedevents}
Sachin Shah, Matthew~A Chan, Haoming Cai, Jingxi Chen, Sakshum Kulshrestha, Chahat~Deep Singh, Yiannis Aloimonos, and Christopher~A Metzler.
\newblock Codedevents: optimal point-spread-function engineering for 3d-tracking with event cameras.
\newblock In \emph{Proceedings of the IEEE/CVF conference on computer vision and pattern recognition}, pages 25265--25275, 2024.

\bibitem[Song et~al.(2021)Song, Sohl-Dickstein, Kingma, Kumar, Ermon, and Poole]{song2020score}
Yang Song, Jascha Sohl-Dickstein, Diederik~P Kingma, Abhishek Kumar, Stefano Ermon, and Ben Poole.
\newblock Score-based generative modeling through stochastic differential equations.
\newblock In \emph{ICLR}, 2021.

\bibitem[Stoffregen et~al.(2020)Stoffregen, Scheerlinck, Scaramuzza, Drummond, Barnes, Kleeman, and Mahony]{stoffregen2020reducing}
Timo Stoffregen, Cedric Scheerlinck, Davide Scaramuzza, Tom Drummond, Nick Barnes, Lindsay Kleeman, and Robert Mahony.
\newblock Reducing the sim-to-real gap for event cameras.
\newblock In \emph{Computer Vision--ECCV 2020: 16th European Conference, Glasgow, UK, August 23--28, 2020, Proceedings, Part XXVII 16}, pages 534--549. Springer, 2020.

\bibitem[Sun et~al.(2023)Sun, Sakaridis, Liang, Sun, Cao, Zhang, Jiang, Wang, and Van~Gool]{sun2023event}
Lei Sun, Christos Sakaridis, Jingyun Liang, Peng Sun, Jiezhang Cao, Kai Zhang, Qi Jiang, Kaiwei Wang, and Luc Van~Gool.
\newblock Event-based frame interpolation with ad-hoc deblurring.
\newblock In \emph{Proceedings of the IEEE/CVF Conference on Computer Vision and Pattern Recognition}, pages 18043--18052, 2023.

\bibitem[Tulyakov et~al.(2021)Tulyakov, Gehrig, Georgoulis, Erbach, Gehrig, Li, and Scaramuzza]{tulyakov2021time}
Stepan Tulyakov, Daniel Gehrig, Stamatios Georgoulis, Julius Erbach, Mathias Gehrig, Yuanyou Li, and Davide Scaramuzza.
\newblock Time lens: Event-based video frame interpolation.
\newblock In \emph{Proceedings of the IEEE/CVF conference on computer vision and pattern recognition}, pages 16155--16164, 2021.

\bibitem[Tulyakov et~al.(2022)Tulyakov, Bochicchio, Gehrig, Georgoulis, Li, and Scaramuzza]{tulyakov2022time}
Stepan Tulyakov, Alfredo Bochicchio, Daniel Gehrig, Stamatios Georgoulis, Yuanyou Li, and Davide Scaramuzza.
\newblock Time lens++: Event-based frame interpolation with parametric non-linear flow and multi-scale fusion.
\newblock In \emph{Proceedings of the IEEE/CVF Conference on Computer Vision and Pattern Recognition}, pages 17755--17764, 2022.

\bibitem[Wang et~al.(2024)Wang, Obukhov, and Schindler]{wang2024consistency}
Tianfu Wang, Anton Obukhov, and Konrad Schindler.
\newblock Consistency\^{} 2: Consistent and fast 3d painting with latent consistency models.
\newblock \emph{arXiv preprint arXiv:2406.11202}, 2024.

\bibitem[Wang et~al.(2004)Wang, Bovik, Sheikh, and Simoncelli]{wang2004image}
Zhou Wang, Alan~C Bovik, Hamid~R Sheikh, and Eero~P Simoncelli.
\newblock Image quality assessment: from error visibility to structural similarity.
\newblock \emph{IEEE transactions on image processing}, 13\penalty0 (4):\penalty0 600--612, 2004.

\bibitem[Wu et~al.(2024)Wu, Tao, Li, Wang, Liu, and Zheng]{Wu_2024_CVPR}
Guangyang Wu, Xin Tao, Changlin Li, Wenyi Wang, Xiaohong Liu, and Qingqing Zheng.
\newblock Perception-oriented video frame interpolation via asymmetric blending.
\newblock In \emph{Proceedings of the IEEE/CVF Conference on Computer Vision and Pattern Recognition (CVPR)}, pages 2753--2762, 2024.

\bibitem[Xing et~al.(2024)Xing, Xia, Zhang, Chen, Yu, Liu, Liu, Wang, Shan, and Wong]{xing2024dynamicrafter}
Jinbo Xing, Menghan Xia, Yong Zhang, Haoxin Chen, Wangbo Yu, Hanyuan Liu, Gongye Liu, Xintao Wang, Ying Shan, and Tien-Tsin Wong.
\newblock Dynamicrafter: Animating open-domain images with video diffusion priors.
\newblock In \emph{European Conference on Computer Vision}, pages 399--417. Springer, 2024.

\bibitem[Yuan et~al.(2024)Yuan, Burner, Wu, Liu, Chen, Aloimonos, and Ferm{\"u}ller]{yuan2024learning}
Dehao Yuan, Levi Burner, Jiayi Wu, Minghui Liu, Jingxi Chen, Yiannis Aloimonos, and Cornelia Ferm{\"u}ller.
\newblock Learning normal flow directly from event neighborhoods.
\newblock \emph{arXiv preprint arXiv:2412.11284}, 2024.

\bibitem[Zhang et~al.(2023{\natexlab{a}})Zhang, Zhu, Wang, Chen, Wu, and Wang]{zhang2023extracting}
Guozhen Zhang, Yuhan Zhu, Haonan Wang, Youxin Chen, Gangshan Wu, and Limin Wang.
\newblock Extracting motion and appearance via inter-frame attention for efficient video frame interpolation.
\newblock In \emph{Proceedings of the IEEE/CVF Conference on Computer Vision and Pattern Recognition}, pages 5682--5692, 2023{\natexlab{a}}.

\bibitem[Zhang et~al.(2023{\natexlab{b}})Zhang, Rao, and Agrawala]{zhang2023adding}
Lvmin Zhang, Anyi Rao, and Maneesh Agrawala.
\newblock Adding conditional control to text-to-image diffusion models.
\newblock In \emph{ICCV}, 2023{\natexlab{b}}.

\bibitem[Zhang et~al.(2018)Zhang, Isola, Efros, Shechtman, and Wang]{zhang2018unreasonable}
Richard Zhang, Phillip Isola, Alexei~A Efros, Eli Shechtman, and Oliver Wang.
\newblock The unreasonable effectiveness of deep features as a perceptual metric.
\newblock In \emph{Proceedings of the IEEE conference on computer vision and pattern recognition}, pages 586--595, 2018.

\bibitem[Zhang and Yu(2022)]{zhang2022unifying}
Xiang Zhang and Lei Yu.
\newblock Unifying motion deblurring and frame interpolation with events.
\newblock In \emph{Proceedings of the IEEE/CVF Conference on Computer Vision and Pattern Recognition}, pages 17765--17774, 2022.

\bibitem[Zhang et~al.(2024)Zhang, Gu, Wang, Wang, Cheng, Zhu, and Zou]{zhang2024mimicmotion}
Yuang Zhang, Jiaxi Gu, Li-Wen Wang, Han Wang, Junqi Cheng, Yuefeng Zhu, and Fangyuan Zou.
\newblock Mimicmotion: High-quality human motion video generation with confidence-aware pose guidance.
\newblock \emph{arXiv preprint arXiv:2406.19680}, 2024.

\bibitem[Zhong et~al.(2024)Zhong, Krishnan, Sun, Qiao, Ma, and Wang]{zhong2024clearer}
Zhihang Zhong, Gurunandan Krishnan, Xiao Sun, Yu Qiao, Sizhuo Ma, and Jian Wang.
\newblock Clearer frames, anytime: Resolving velocity ambiguity in video frame interpolation.
\newblock In \emph{European Conference on Computer Vision}, pages 346--363. Springer, 2024.

\bibitem[Zhu et~al.(2018)Zhu, Yuan, Chaney, and Daniilidis]{zhu2018ev}
Alex~Zihao Zhu, Liangzhe Yuan, Kenneth Chaney, and Kostas Daniilidis.
\newblock Ev-flownet: Self-supervised optical flow estimation for event-based cameras.
\newblock \emph{arXiv preprint arXiv:1802.06898}, 2018.

\bibitem[Zhu et~al.(2019)Zhu, Yuan, Chaney, and Daniilidis]{zhu2019unsupervised}
Alex~Zihao Zhu, Liangzhe Yuan, Kenneth Chaney, and Kostas Daniilidis.
\newblock Unsupervised event-based learning of optical flow, depth, and egomotion.
\newblock In \emph{Proceedings of the IEEE/CVF Conference on Computer Vision and Pattern Recognition}, pages 989--997, 2019.

\end{thebibliography}
}

% WARNING: do not forget to delete the supplementary pages from your submission 
% \documentclass{article}
% \begin{document}

\clearpage
\maketitlesupplementary

\tableofcontents

\section{ Video Results}
Please refer to our project page: \url{https://vdm-evfi.github.io/} for video results, which clearly demonstrate that our reconstructions provide superior consistency and generalization compared to other baselines.

\section{ Video Generation Task Results}
As explained in the main paper, our method supports Event-based Video Generation, an extrapolation task that relies on only one frame (start or end) and events, unlike interpolation, which uses both frames. This constraint in video generation leads to error accumulation in the generated video, as shown in the last video of the website.
We present a comparison of video generation and interpolation results on the BS-ERGB dataset, for video generation, only the left-end frame next to skip frames and corresponding events are used to generate the skipped frames, as shown in Table \ref{tab:compareGeneration}, relying on information from only one side instead of both start and end frames causes the PSNR and SSIM metrics to drop significantly compared to the interpolation results.
% \vspace{-2mm}

\begin{table}[tbh!]
    \centering
    \resizebox{0.9\linewidth}{!}{
    \begin{tabular}{@{}ccccccccccc@{}}
    \midrule
    \midrule
    \multirow{2}{*}{\textbf{Task}}  &  \multicolumn{3}{|c|}{\textbf{BS-ERGB (3 skips)}}  \\ 
    \cmidrule(r){2-4}
     &  PSNR $\uparrow$ & SSIM $\uparrow$  & LPIPS $\downarrow$  \\ 
        \midrule
        Video Generation   &  25.79
                & 0.84   & 0.11  \\
        Video Interpolation   & 27.74  & 0.88 & 0.12  \\
        \midrule
        \midrule
    \end{tabular}
    }
    \vspace{-2mm}
    \captionsetup{font=small}
    \caption{Comparison of our pipeline's performance on Video Generation and Video Interpolation tasks on the BS-ERGB dataset.}
    \label{tab:compareGeneration}
    \vspace{-10pt}
\end{table}
% \vspace{-3mm}

\begin{figure}[tbh!]
  \centering
  \includegraphics[width=1\linewidth]{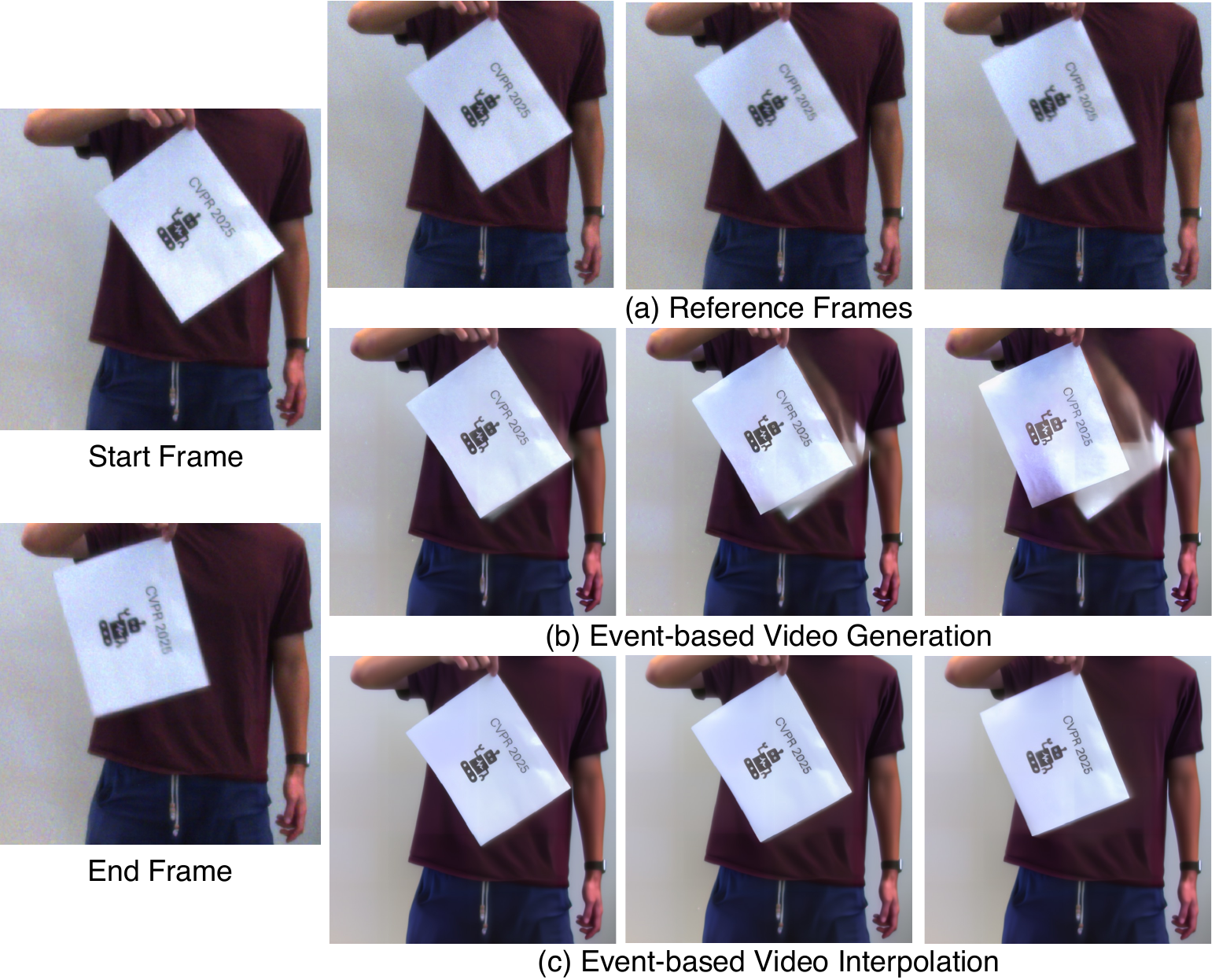}
   \caption{An example illustrating the difference between event-based video generation, which relies solely on the start frame, and event-based video interpolation, which uses both start and end frames to infer the interpolated frames. We present the 3rd, 6th, and 9th interpolated frames for the 11 skips interpolation between start and end frames on Clear-Motion test sequences. In the video generation scenario, the model hallucinates the occluded parts behind the paper due to the lack of information in the start frame. In contrast, video interpolation avoids hallucination as the end frame provides the necessary information.}
   \label{fig:compare_gen}
\end{figure}

\begin{figure}[tbh!]
  \centering
  \includegraphics[width=1\linewidth]{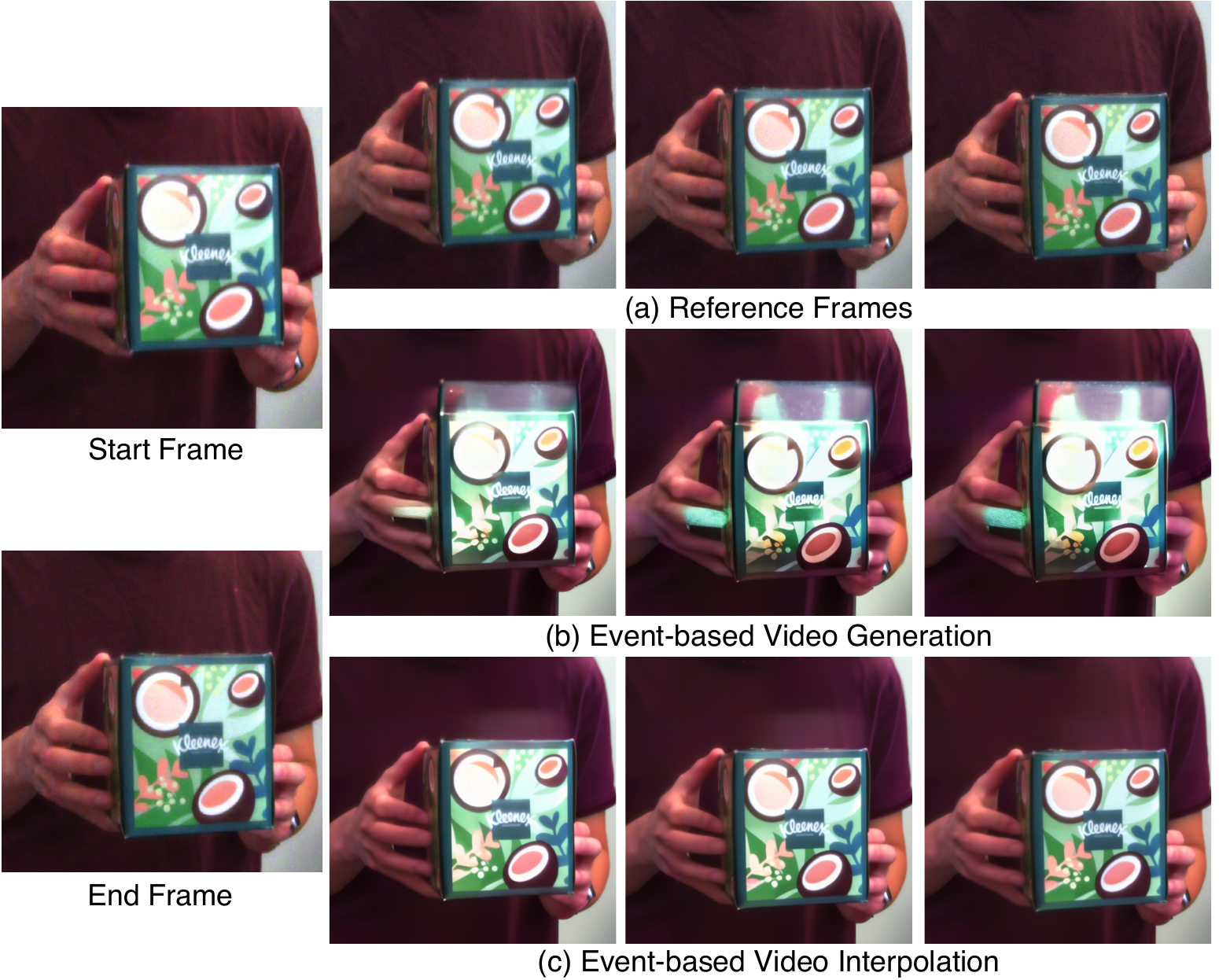}
   \caption{An example illustrating the difference in video consistency between event-based video generation and interpolation for 11 skips between start and end frames on Clear-Motion test sequences, we present the 5th, 8th, and 11th interpolated frames. In the video generation task, color errors on the finger accumulate over time, with the 11th frame (the final interpolated frame) failing to align with the information in the end frame. In contrast, interpolation reduces error accumulation and ensures consistency by leveraging information from both the start and end frames. }
   \label{fig:compare_gen_consist}
\end{figure}

We also show a qualitative comparison in Figure \ref{fig:compare_gen}, as we can see because the video generation task is a extrapolation task that only uses the information from one side (start or end frame) of frames and events, it cannot avoid hallucination for the occluded/missing parts that are not present in the condition image. In contrast, the video generation task can mitigate the hallucination well by using complementary information from both frames. 
% \vspace{mm}
In addition to hallucination, relying on frame information from only one side leads to error accumulation during generation, as shown in Figure \ref{fig:compare_gen_consist}, the results in video generation (incorrect color accumulation on the finger) are inconsistent with the information provided by the frame on the other side. In contrast, interpolation ensures the reconstructed video remains consistent with the information from both frames. The comparison of video consistency is best observed in the last video of the website, which compares event-based video generation with event-based video interpolation.

%%%%%%%%% BODY TEXT - ENTER YOUR RESPONSE BELOW
\section{ Clear-Motion Test Sequences}
To robustly evaluate the zero-shot generalization performance of all models on unseen real-world event-based video frame interpolation scenarios, we collected the Clear-Motion Test Sequences solely for testing purposes.
%-------------------------------------------------------------------------

\subsection{ Event-RGB Aligned Video Capture Setup}
\begin{figure}[tbh!]
  \centering
  \includegraphics[width=1\linewidth]{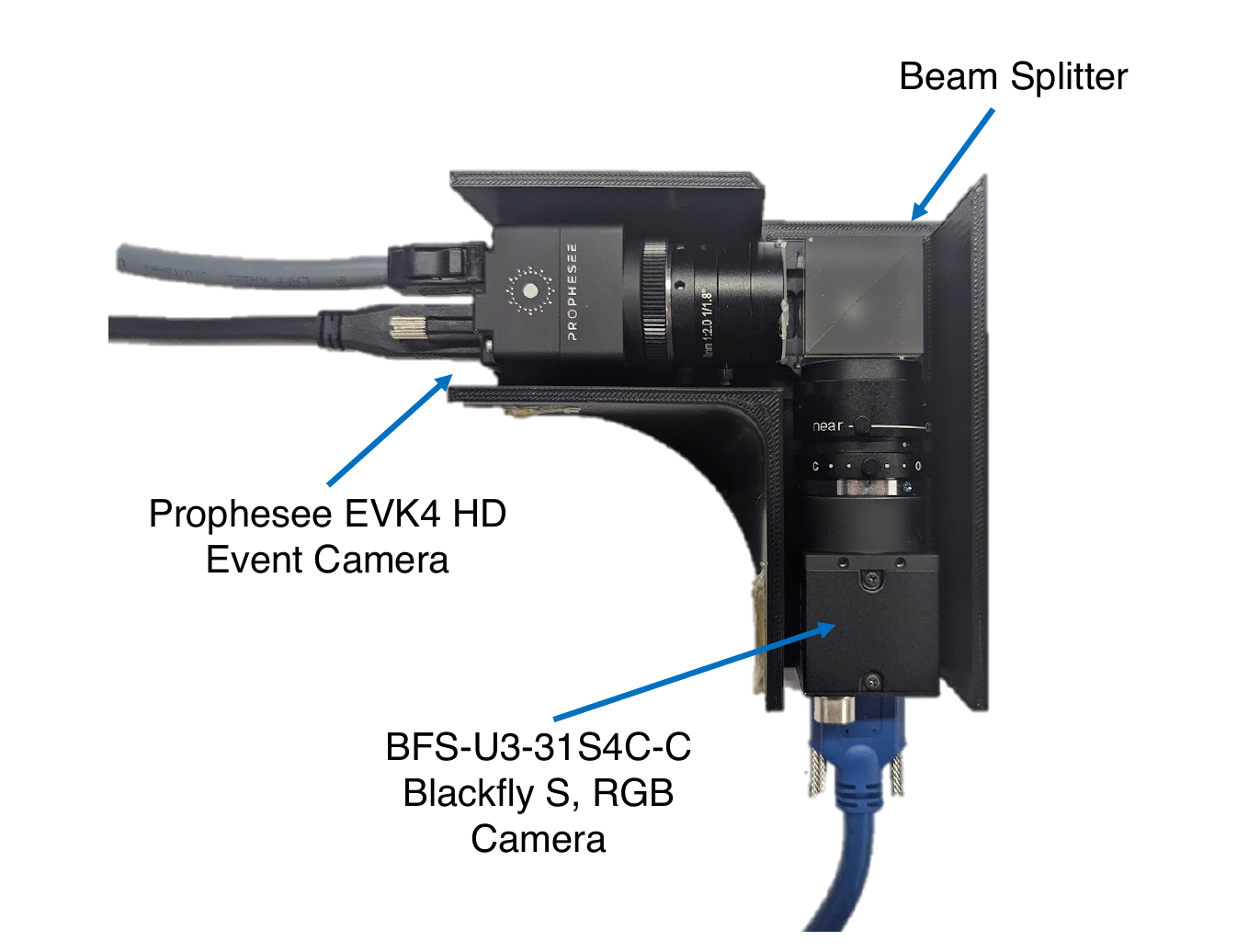}
   \caption{An illustration of our capture setup for the Clear-Motion test sequences of Event-RGB aligned video sequences. The setup consists of three main components: an event camera, an RGB camera, and a beam splitter to align the field of view for both cameras.}
   \label{fig:capture_setup}
\end{figure}

In this section, we will present the capture setup for our event-rgb aligned video sequences, as shown in Figure \ref{fig:capture_setup}.

We use the Prophesee EVK4 HD as our event camera, offering a capture resolution of 1280 $\times$ 720. For the RGB camera, we use the BFS-U3-31S4C-C Blackfly S, which provides a resolution of 2048 $\times$ 1536 and supports up to 55 frames per second (fps). To align the field of view for both cameras, we utilize the Thorlabs CCM1-BS013 30 mm Cage Cube-Mounted Non-Polarizing Beam Splitter. Additionally, we perform spatial and temporal alignment to synchronize the events with the captured RGB frames. After the spatial alighment, our final captured RGB frames aligned with event are of resolution 940 $\times$ 720 and 40 fps.

\subsection{ Details of Data Sequence}
\begin{figure}[tbh!]
  \centering
  \includegraphics[width=1\linewidth]{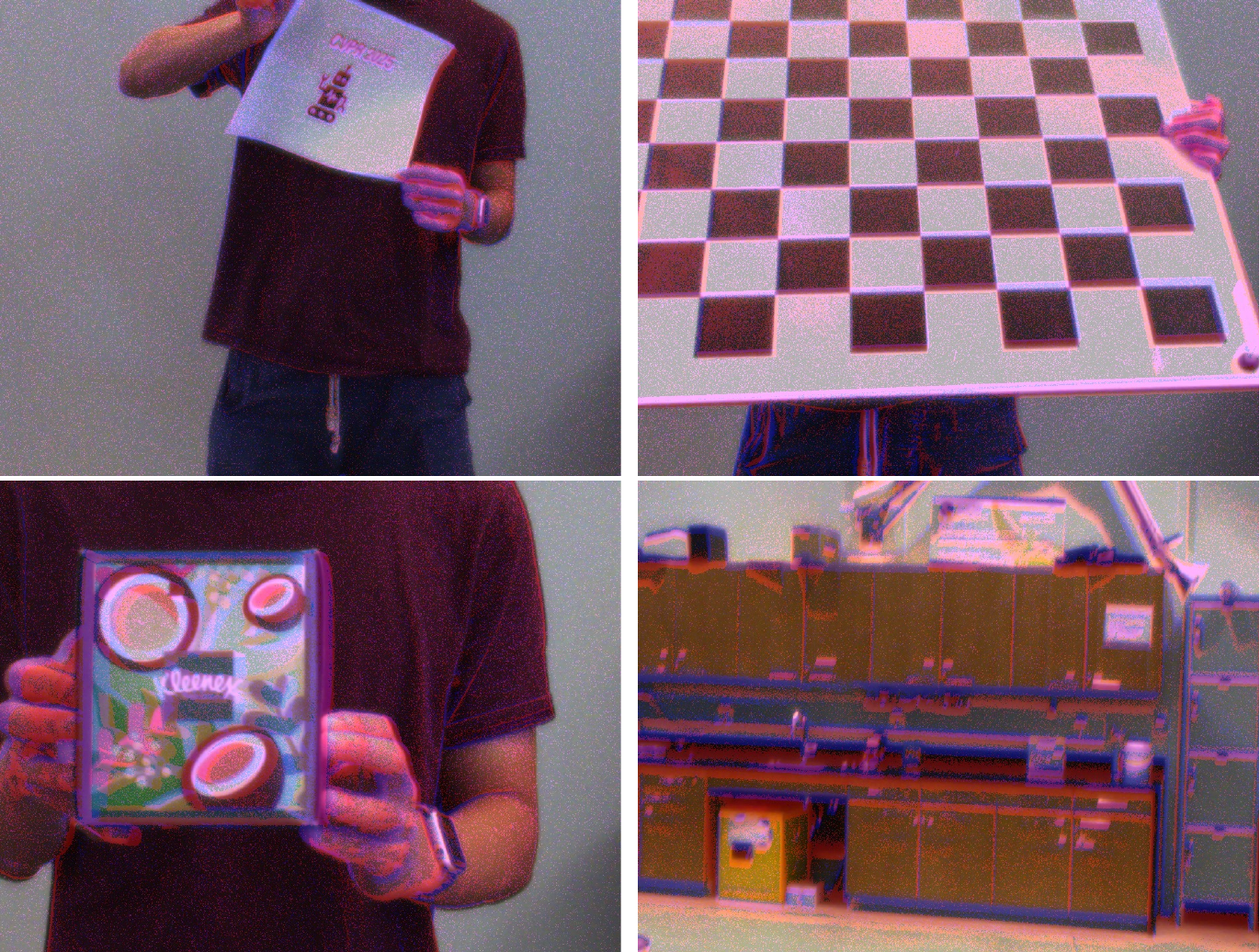}
   \caption{An illustration of example frames overlaid with events from Clear-Motion test sequences.}
   \label{fig:dataset}
\end{figure}

The Clear-Motion test sequences are designed to include clear and large motions, encompassing both camera and object movements, as well as objects and motion patterns distinct from those found in most existing real-world Event-based Video Frame Interpolation (EVFI) datasets \cite{tulyakov2021time, Kim_2023_CVPR, stoffregen2020reducing}. This setup enables a straightforward evaluation of the generalization and consistency of various video frame interpolation methods. Table \ref{tab:dataset_details} provides details of our collected test sequences, including explanations for each sequence. Category (i) represents sequences with object motion, while Category (ii) represents sequences with camera motion. We also include a Figure \ref{fig:dataset} to show some example data in our test sequences.

\begin{table*}[tbh!]
    \centering
    \resizebox{0.8\linewidth}{!}{
    \begin{tabular}{@{}cccccc@{}} % Four columns
    \midrule
    \midrule
    & \makecell{\textbf{Sequence}} & \makecell{\textbf{\#Frames}} & \makecell{\textbf{Explanation}} & \makecell{\textbf{Category}} \\
    \midrule
     & Paper\_Shifting  & 200 & The translational and 3D rotational motion of a paper with a simple texture &  (i) \\
     & Paper\_Waving  & 200 & The waving and 3D rotational motion of a paper with a simple texture &  (i) \\
     & Paper\_Deforming  & 200 & The deformation motion of a paper with a simple texture &  (i) \\
     & Camera\_Far  & 200 & The moving cameras capturing distant objects &  (ii) \\
     & Camera\_Close  & 200 & The moving cameras capturing nearby objects &  (ii) \\
     & Checkerboard\_Planar  & 200 & The planar translation and rotation of a nearby dense checkerboard &  (i) \\
     & Checkerboard\_Depth  & 200 & The motion of a checkerboard along the depth direction &  (i) \\
     & Checkerboard\_3D  & 200 & The 3D translation and rotation of a nearby dense checkerboard &  (i) \\
     & Texture\_Box  & 200 & The translation and rotation of a nearby highly textured box &  (i) \\
    \midrule
    \midrule
    \end{tabular}}

    \captionsetup{font=small}
    \caption{A detailed description of our collected Clear-Motion test sequences, with sequence name, number of frames and explanation of each sequence. Category (i) includes sequences with object motion, while Category (ii) includes sequences with camera motion. }
    \label{tab:dataset_details}

    \vspace{-3mm}
\end{table*}

%------------------------------------------------------------------------
\section{ The Impact of Input Upsampling}

As discussed in the main paper, to mitigate the loss of appearance and motion control accuracy caused by the conversion between downsampled latent space and pixel space in Latent Diffusion Models (LDM) \cite{blattmann2023stable}, we employ test-time optimization in the Per-tile Denoising and Fusion process. This involves first upsampling the input image and event representations by a specified factor and then breaking them into fixed-size overlapping tiles before feeding them into the video diffusion process. The performance comparison across different upsampling factors is shown in Table \ref{tab:upsample}. As the upsampling factor increases from 1 to 2, our model's performance improves significantly, with PSNR increasing by approximately 3 dB, SSIM by 0.11, and LPIPS decreasing by 0.03. These results demonstrate the effectiveness of upsampling in the Per-tile Denoising and Fusion process, enhancing both the details in reconstructed frames and the accuracy of event-based motion control.

\begin{table}[tbh!]
    \centering
    \resizebox{0.8\linewidth}{!}{
    \begin{tabular}{@{}ccccccccccc@{}}
    \midrule
    \midrule
    \multirow{2}{*}{\textbf{Method}}  &  \multicolumn{3}{|c|}{\textbf{BS-ERGB (3 skips)}}  \\ 
    \cmidrule(r){2-4}
     &  PSNR $\uparrow$ & SSIM $\uparrow$  & LPIPS $\downarrow$  \\ 
        \midrule
        Ours\_1  &  24.82
                & 0.77   & 0.15 \\
        Ours\_1.5  &  25.86
                & 0.82   & 0.16  \\
        Ours\_2   & 27.74  & 0.88 & 0.12  \\
        \midrule
        \midrule
    \end{tabular}
    }
    \vspace{-2mm}
    \captionsetup{font=small}
    \caption{Comparison of the impact of different upsampling factors \{1, 1.5, 2\} on our model's performance on the BS-ERGB dataset. We use 512 $\times$ 320 overlapping tiles with a overlapping ratio 0.1 for the input image and event representations.}
    \label{tab:upsample}
\end{table}

\begin{figure}[tbh!]
  \centering
  \includegraphics[width=1\linewidth]{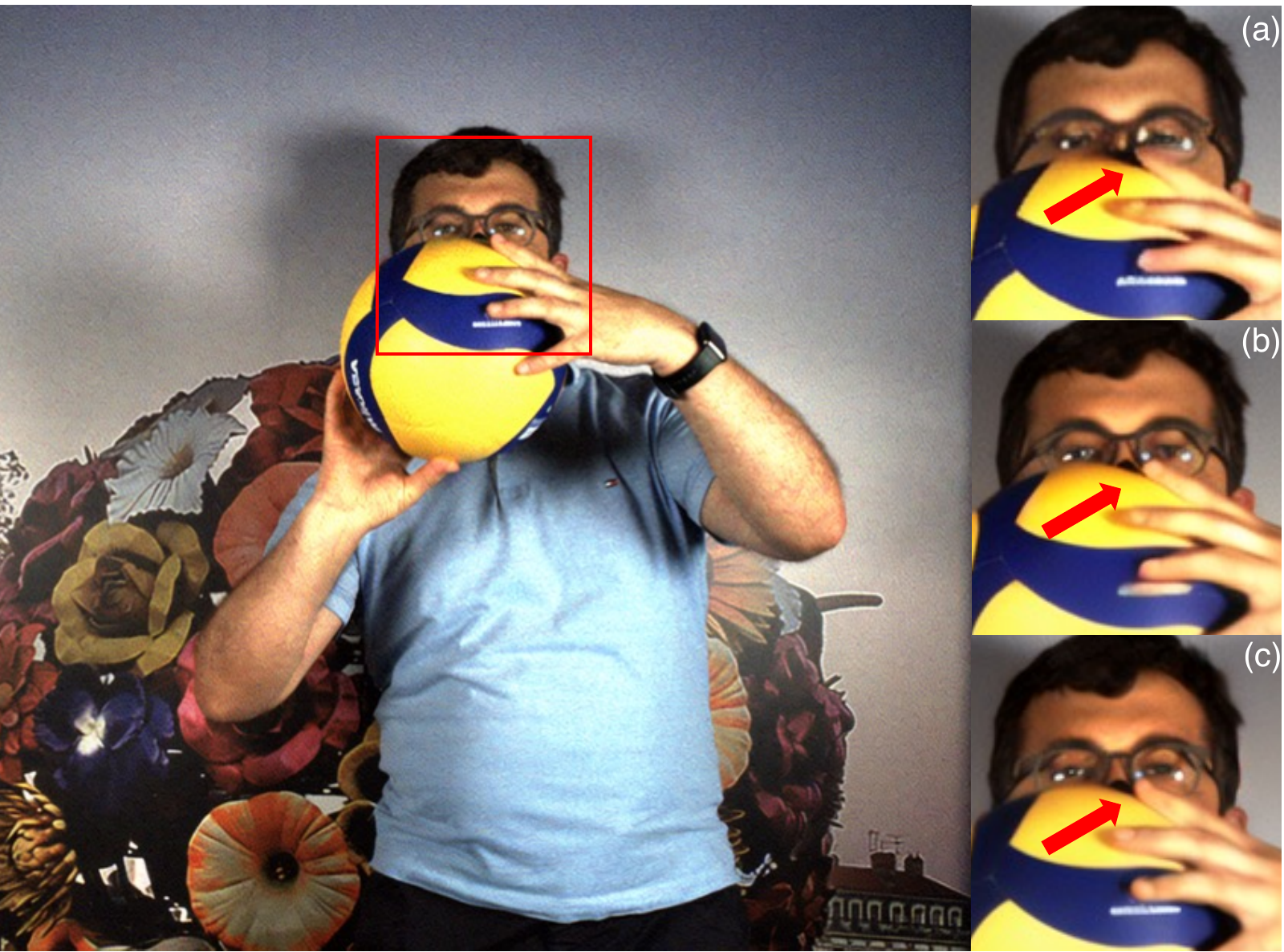}
   \caption{An illustration showcasing the qualitative impact of input upsampling factors. The leftmost image is the reference frame. (a) shows the interpolated result with an upsampling factor of 1, (b) with a factor of 1.5, and (c) with a factor of 2. As the upsampling factor increases from 1 to 2, the details on the human eyes and fingers, highlighted by red arrows, improve significantly.}
   \label{fig:upsample_supp}
\end{figure}

To qualitatively assess the effect of upsampling, Figure \ref{fig:upsample_supp} shows that as the upsampling factor increases from 1 to 2, details on the human eyes and fingers (e.g., nails and textures) improve significantly. Both quantitative and qualitative results highlight the effectiveness of upsampling in the Per-tile Denoising and Fusion process, enhancing the realism and event-based motion control accuracy of video interpolation results.

% \subsection{Video Generation Only Performance }

% \begin{itemize}
%     \item Table on datasets
%     \item One Figure
% \end{itemize}

% \begin{figure}[tbh!]
%   \centering
%   \includegraphics[width=1\linewidth]{sec/Figures/Supp/exp_real_supp_gen.pdf}
%    \caption{Example of caption.  It is set in Roman so that mathematics
%    (always set in Roman: $B \sin A = A \sin B$) may be included without an
%    ugly clash.}
%    \label{fig:onecol}
% \end{figure}

% %-------------------------------------------------------------------------
% \subsection{The Impact of Upsampling and Overlapping Factors}
% \begin{itemize}
%     \item Table on datasets
%     \item One Figure
% \end{itemize}

\section{ More Visual Results}
\begin{figure*}[tbh!]
  \centering
  \includegraphics[width=1\linewidth]{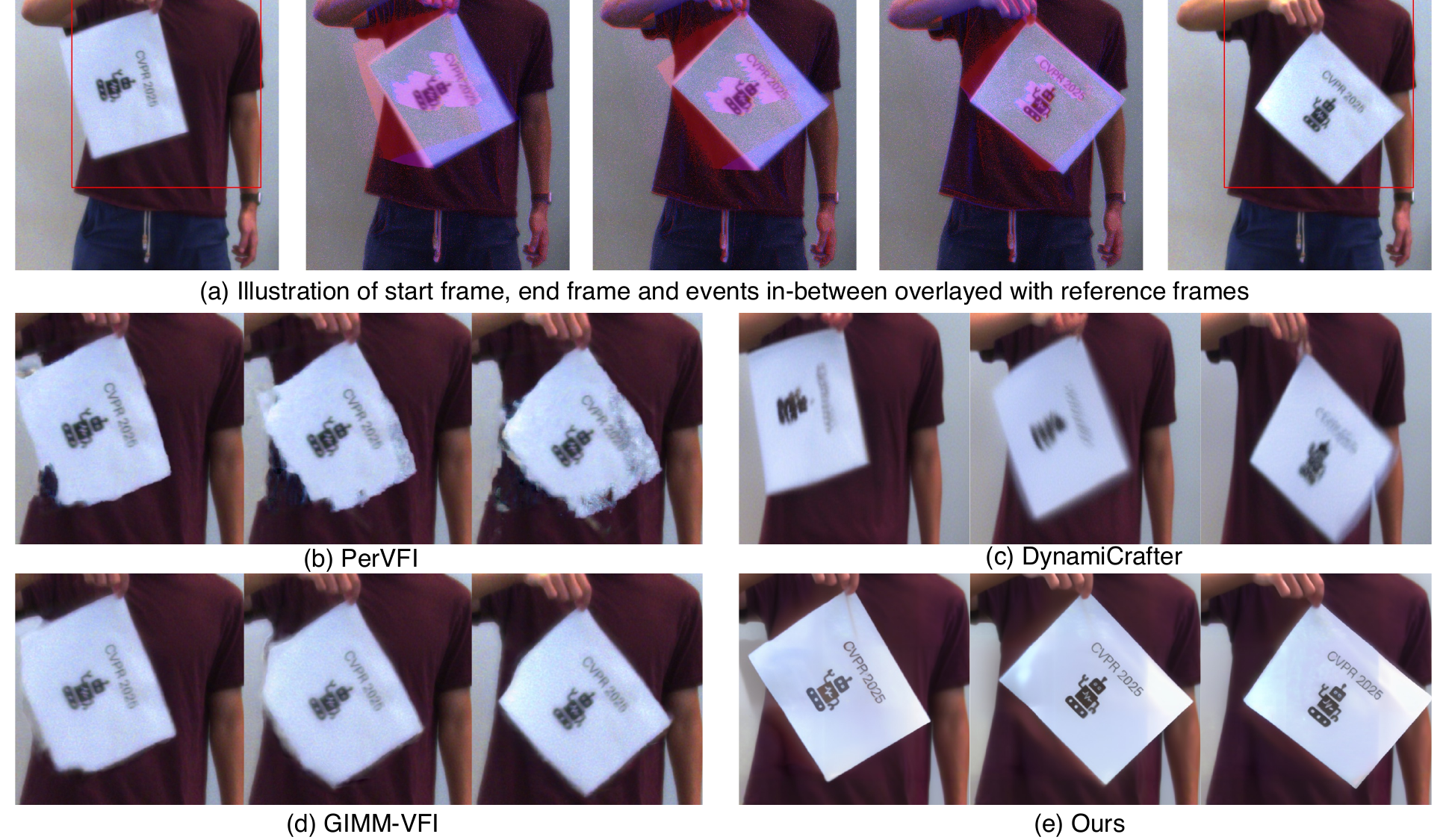}
   \caption{Additional baseline results on the Clear-Motion sequence Paper\_Waving.}
   \label{fig:rebuttal_1}
\end{figure*}

\begin{figure*}[tbh!]
  \centering
  \includegraphics[width=1\linewidth]{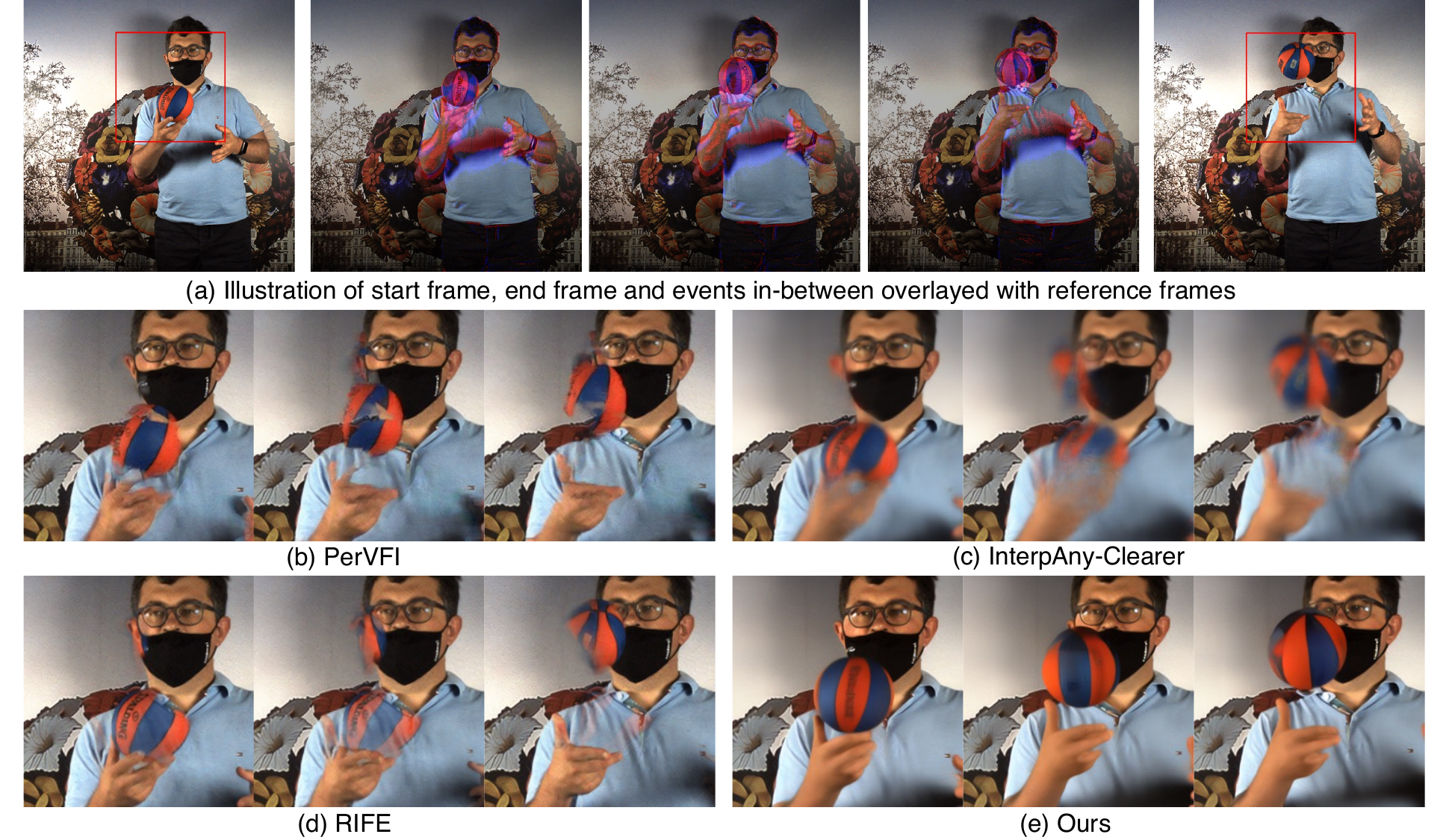}
   \caption{Additional baseline results on the BS-ERGB sequence as presented in the main paper.}
   \label{fig:rebuttal_2}
\end{figure*}

\begin{figure*}[tbh!]
  \centering
  \includegraphics[width=1\linewidth]{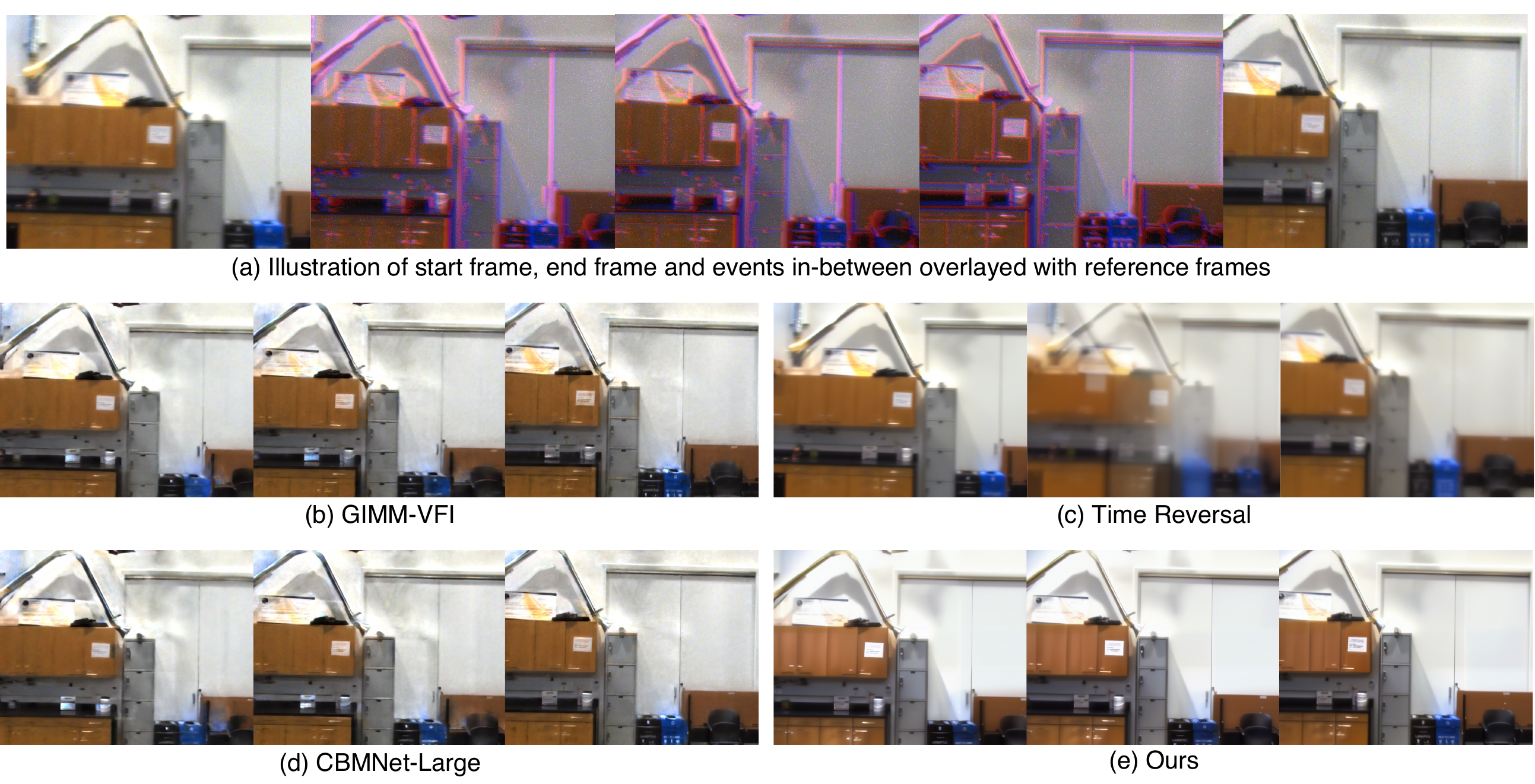}
   \caption{An illustration showcasing the qualitative comparison on the Clear-Motion sequence Camera\_Far, which involves large camera motion capturing distant objects, with 11 skips between the start and end frames. We present the interpolated 4th, 7th, and 10th frames. (Zoom in for the best viewing experience)}
   \label{fig:exp_supp_1}
\end{figure*}

\begin{figure*}[tbh!]
  \centering
  \includegraphics[width=1\linewidth]{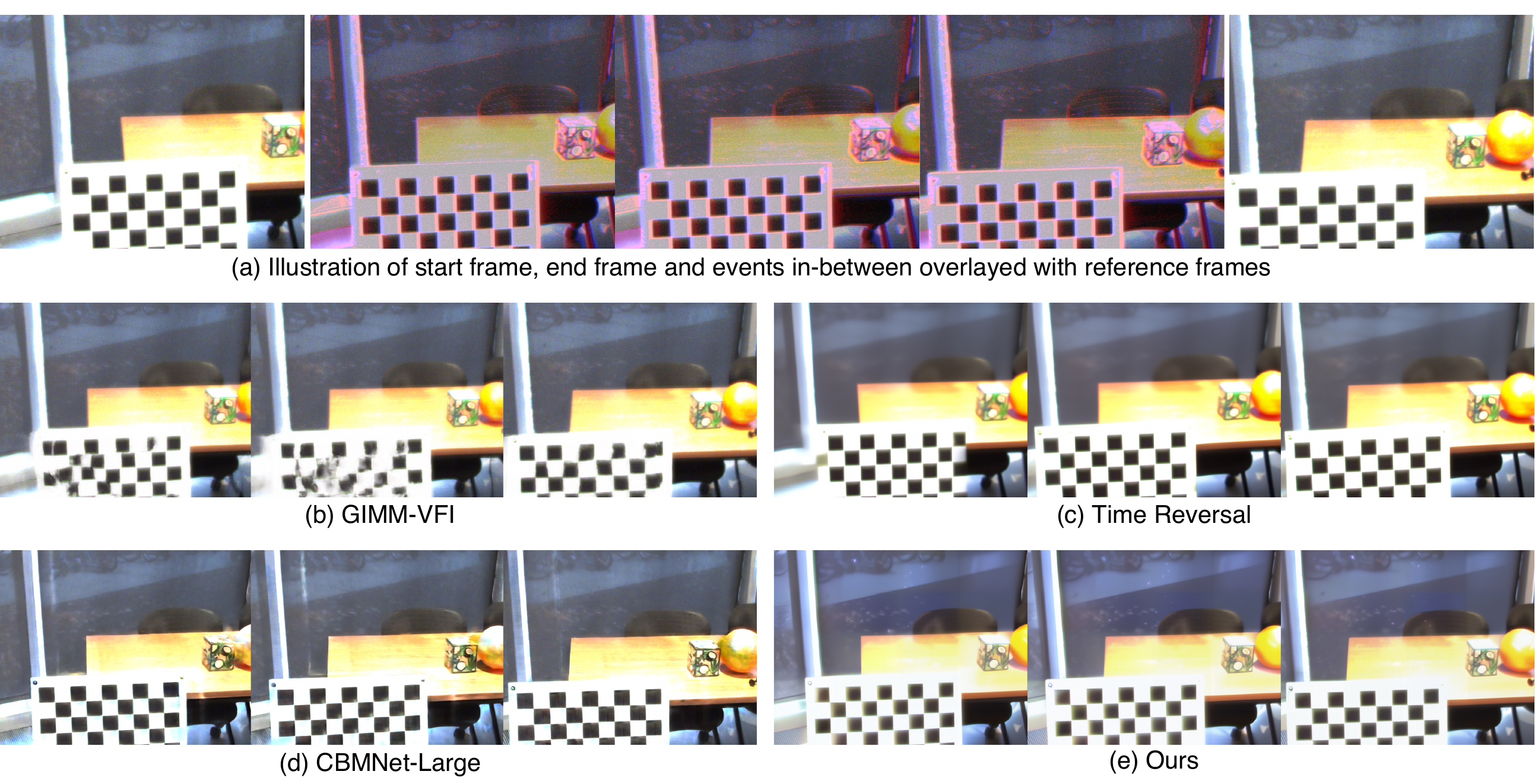}
   \caption{An illustration showcasing the qualitative comparison on the Clear-Motion sequence Camera\_Close, which involves large camera motion capturing nearby objects, with 11 skips between the start and end frames. We present the interpolated 4th, 7th, and 10th frames. (Zoom in for the best viewing experience)}
   \label{fig:exp_supp_2}
\end{figure*}

\begin{figure*}[tbh!]
  \centering
  \includegraphics[width=1\linewidth]{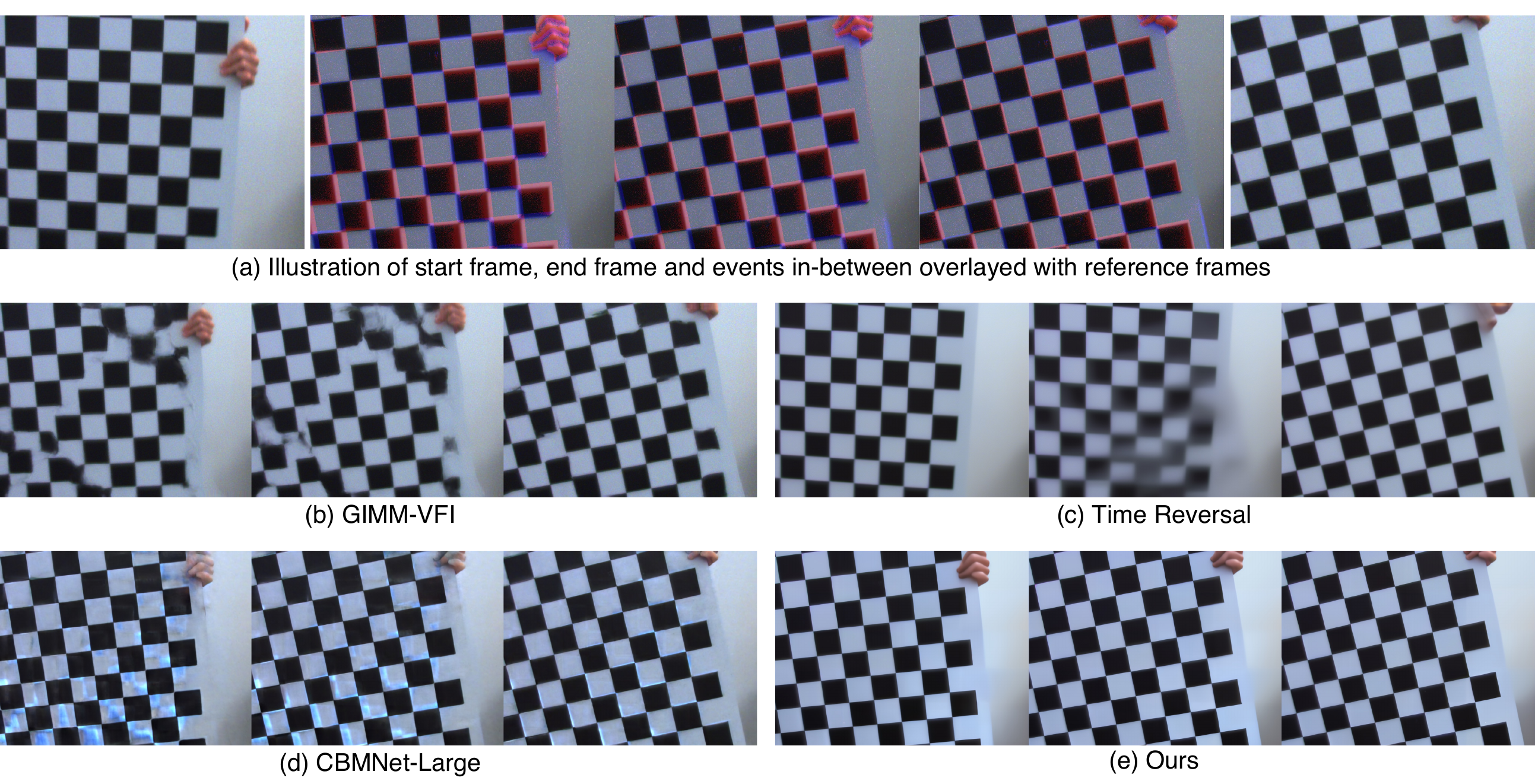}
   \caption{An illustration showcasing the qualitative comparison on the Clear-Motion sequence Checkerboard\_Planar, which involves large planar motion of a nearby checkerboard, with 11 skips between the start and end frames. We present the interpolated 4th, 7th, and 10th frames. (Zoom in for the best viewing experience)}
   \label{fig:exp_supp_3}
\end{figure*}

\begin{figure*}[tbh!]
  \centering
  \includegraphics[width=1\linewidth]{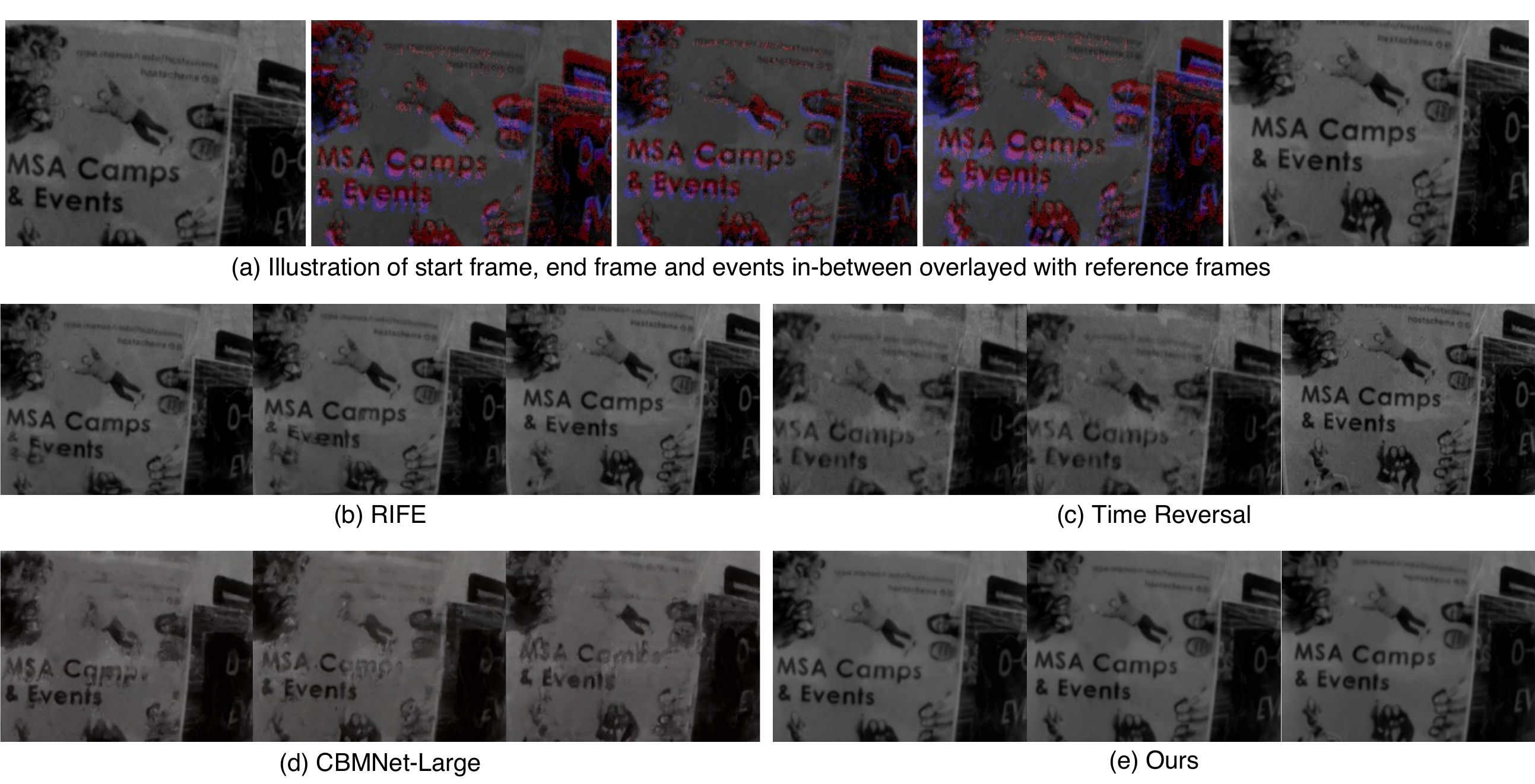}
   \caption{An illustration showcasing the qualitative comparison on the HQF dataset for the sequence poster\_pillar\_1, involving moving cameras capturing nearby posters, with 3 skips between the start and end frames. All interpolated frames are presented. (Zoom in for the best viewing experience)}
   \label{fig:exp_supp_4}
\end{figure*}

In this section, we provide additional visual results showcasing qualitative comparisons between our method and the baselines, as shown in Figures \ref{fig:rebuttal_1}, \ref{fig:rebuttal_2}, \ref{fig:exp_supp_1}, \ref{fig:exp_supp_2}, \ref{fig:exp_supp_3}, and \ref{fig:exp_supp_4}.

\section{ Additional Implementation Details}
In this section, we provide additional implementation details. The pre-trained video diffusion model we used is Stable Video Diffusion \cite{blattmann2023stable} for 14-frame image-to-video generation. We trained our model with an effective batch size of 64, using a batch size of 4 per GPU and a gradient accumulation factor of 16. Training was conducted solely on the BS-ERGB dataset, and the model was tested on other unseen datasets without fine-tuning. All training was performed on 4 NVIDIA RTX A6000 GPUs, each with 50GB of memory.
For training, we use the AdamW optimizer \cite{loshchilov2017fixing} with a learning rate of $5 \times 10^{-5}$ and parameters $\beta_{1} = 0.9$, $\beta_{2} = 0.999$, $\epsilon =1 \times 10^{-8}$, and a weight decay of $1 \times 10^{-2}$. Our model is trained on the BS-ERGB dataset for 72 hours to denoise noisy video latents for 3 and 11 skipped frames. All testing and inference on unseen data or datasets are conducted without fine-tuning, using the checkpoint trained on the BS-ERGB dataset.

During training and testing, the input to our adapted video diffusion model consists of 512 × 320 size tiles. Our model is trained solely to denoise/generate video latents using start frames and forward-time events. For testing and inference, we use an overlapping ratio of 0.1 for overlapping tiles and set the number of denoising steps in the video diffusion process to 25. The Per-tile Denoising and Fusion (for high-resolution frame reconstruction and event-based motion control) and Two-side Fusion (for converting video generation to interpolation) are both test-time optimization processes that do not require additional training. 

\section{Model Run Time, Memory, and Parameter Comparison}
 Table \ref{tab:compare_model} reports testing results for all models run on a single NVIDIA RTX 4090 GPU. Each method generated $1024 \times 576$ frames with run time averaged over 16 frames. VDM based methods (Time-Reversal, DynamiCrafter, and Ours) are more memory-intensive and time-consuming than other methods. 

 \begin{table}[t!]
    \centering
    \resizebox{1.0\linewidth}{!}{
    \begin{tabular}{@{}cccccccc@{}} % Four columns
    \midrule
    \midrule
    & \makecell{\textbf{Method}} & \makecell{\textbf{Run Time (s)}} & \makecell{\textbf{Memory Usage (GB)}}  & \makecell{\textbf{\#Parameters (M)}}\\
    \midrule
     & RIFE & 0.5 & 0.9 & 9.8 \\
     & CBMNet-Large & 41.7 & 17.8 & 22.2\\
     & Time-Reversal & 62.6 & 21.5 & 1524.6 \\
     & PerVFI  & 9.3 & 5.2 & 13.9 \\
     & InterpAny-Clearer  & 0.5 & 1.1 & 10.7 \\
     & DynamiCrafter  & 92.0 & 18.3 & 1438.9 \\
     & EMA-VFI  & 1.8 & 4.8 & 65.7 \\
     & GIMM-VFI  & 3.0 & 9.5 & 19.8 \\
     & {Ours}  & {200.1} & {17.2} & {2206.8} \\
    \midrule
    \midrule
    \end{tabular}
    }

    \captionsetup{font=small}
    \vspace{-9pt}
    \caption{Model Run Time, Memory, and Parameter comparison. }
    \vspace{-15pt}
    \label{tab:compare_model}
\end{table}

% \newpage

% \input{Supplementary}

\end{document}